\documentclass[journal]{IEEEtran}
\usepackage{cite}

\ifCLASSINFOpdf
  \usepackage[pdftex]{graphicx}
\else
  \usepackage[dvips]{graphicx}
\fi
\ifCLASSOPTIONcompsoc
  \usepackage[caption=false,font=normalsize,labelfont=sf,textfont=sf]{subfig}
\else
 \usepackage[caption=false,font=footnotesize]{subfig}
\fi

\ifCLASSOPTIONcaptionsoff
  \usepackage[nomarkers]{endfloat}
 \let\MYoriglatexcaption\caption
 \renewcommand{\caption}[2][\relax]{\MYoriglatexcaption[#2]{#2}}
\fi

\usepackage{color}
\usepackage[table]{xcolor}
\definecolor{lightgray}{gray}{0.9}
\usepackage{multirow}

\begin{document}
%
\title{Going Deeper with Contextual CNN for Hyperspectral Image Classification}

%

\author{Hyungtae~Lee,~\IEEEmembership{Member,~IEEE,}
        and~Heesung~Kwon,~\IEEEmembership{Senior Member,~IEEE}
\thanks{Manuscript received October 05, 2016; revised April 13, 2017.}
\thanks{Hyungtae Lee is with Booz Allen Hamilton Inc., McLean, VA, 22102 USA (e-mail: lee\_hyungtae@bah.com).}
\thanks{Heesung Kwon is with the Image processing branch, the Sensors \& Electron Devices Directorate (SEDD), Army Research Laboratory, Adalphi, MD, 20783 USA (e-mail: heesung.kwon.civ@mail.mil).}}
\maketitle

\begin{abstract}

In this paper, we describe a novel deep convolutional neural network (CNN) that is deeper and wider than other existing deep networks for hyperspectral image classification.  Unlike current state-of-the-art approaches in CNN-based hyperspectral image classification, the proposed network, called contextual deep CNN, can optimally explore local contextual interactions by jointly exploiting local spatio-spectral relationships of neighboring individual pixel vectors.  The joint exploitation of the spatio-spectral information is achieved by a multi-scale convolutional filter bank used as an initial component of the proposed CNN pipeline.  The initial spatial and spectral feature maps obtained from the multi-scale filter bank are then combined together to form a joint spatio-spectral feature map.  The joint feature map representing rich spectral and spatial properties of the hyperspectral image is then fed through a fully convolutional network that eventually predicts the corresponding label of each pixel vector.  The proposed approach is tested on three benchmark datasets: the Indian Pines dataset, the Salinas dataset and the University of Pavia dataset.  Performance comparison shows enhanced classification performance of the proposed approach over the current state-of-the-art on the three datasets.

\end{abstract}


%
\IEEEpeerreviewmaketitle

\section{Introduction}

\IEEEPARstart{R}{ecently,} deep convolutional neural networks (DCNN) have been extensively used for a wide range of visual perception tasks, such as object detection/classification, action/activity recognition, etc.  Behind the remarkable success of DCNN on image/video anlaytics are its unique capabilities of extracting underlying nonlinear structures of image data as well as discerning the categories of semantic data contents by jointly optimizing parameters of multiple layers together.

\begin{figure}[t]
\begin{center}
\begin{tabular}{c}
\subfloat[Residual learning]{
\includegraphics[trim = 15mm -10mm 5mm 5mm,width=0.45\linewidth]{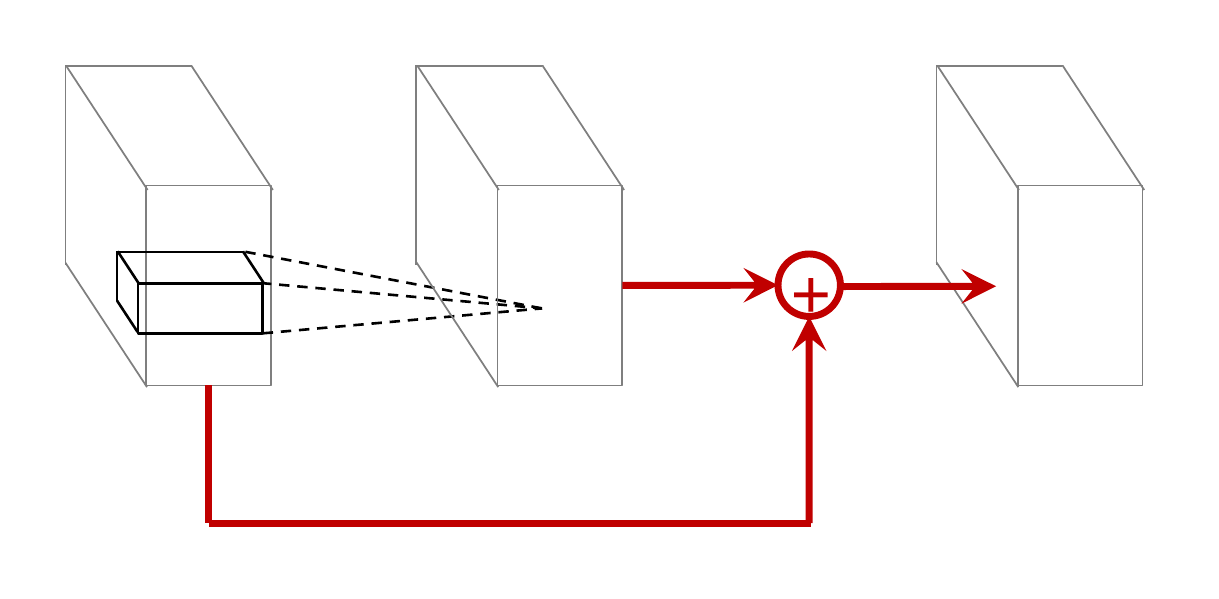}
\label{fig:residual_learning}}
\subfloat[Multi-scale filter bank]{
\includegraphics[trim = 10mm 0mm 5mm 5mm,width=0.45\linewidth]{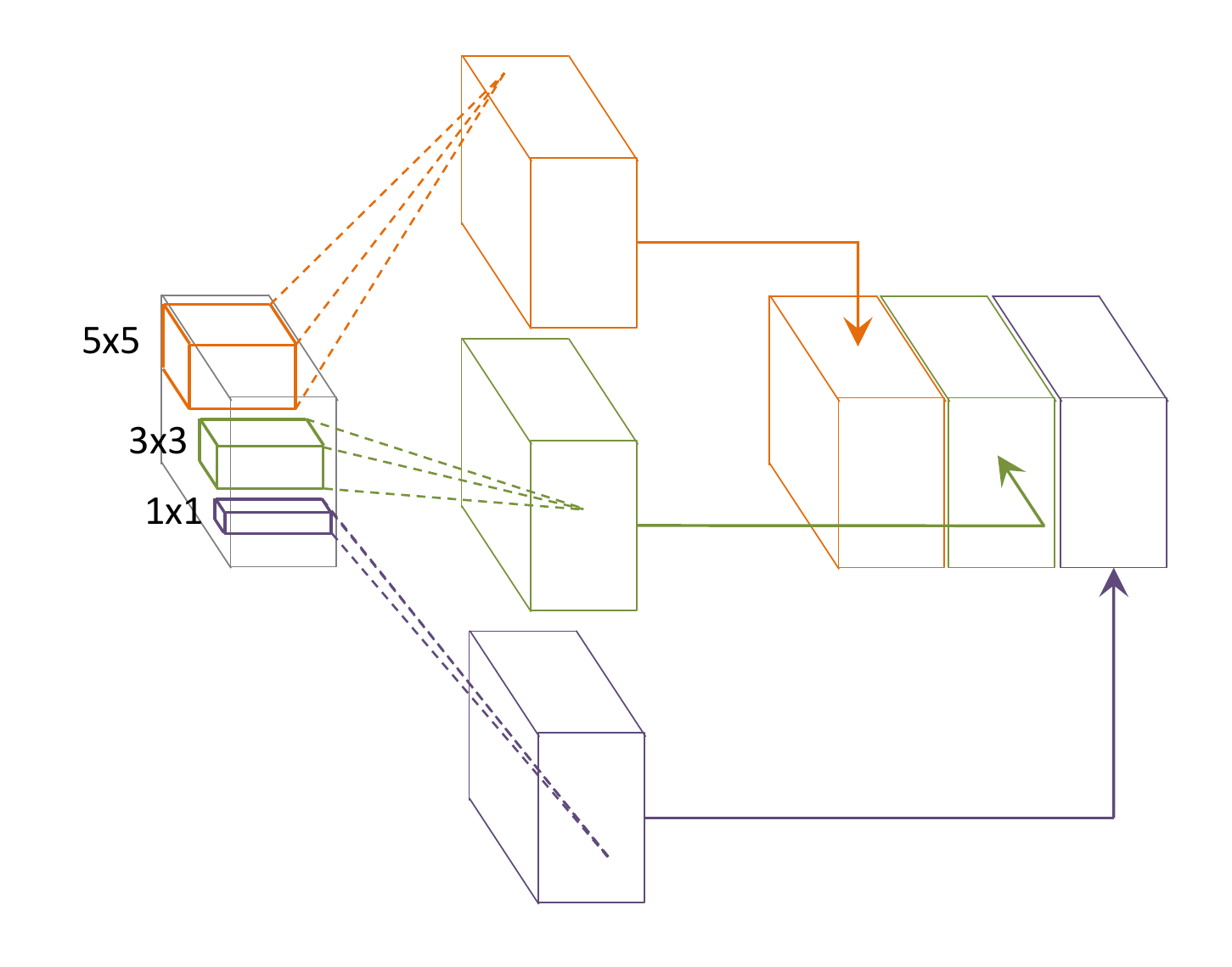}
\label{fig:filter_bank}
}
\\
\subfloat[Fully Convolutional Network (FCN)]{
\includegraphics[trim = 15mm 5mm 5mm -5mm,width=\linewidth]{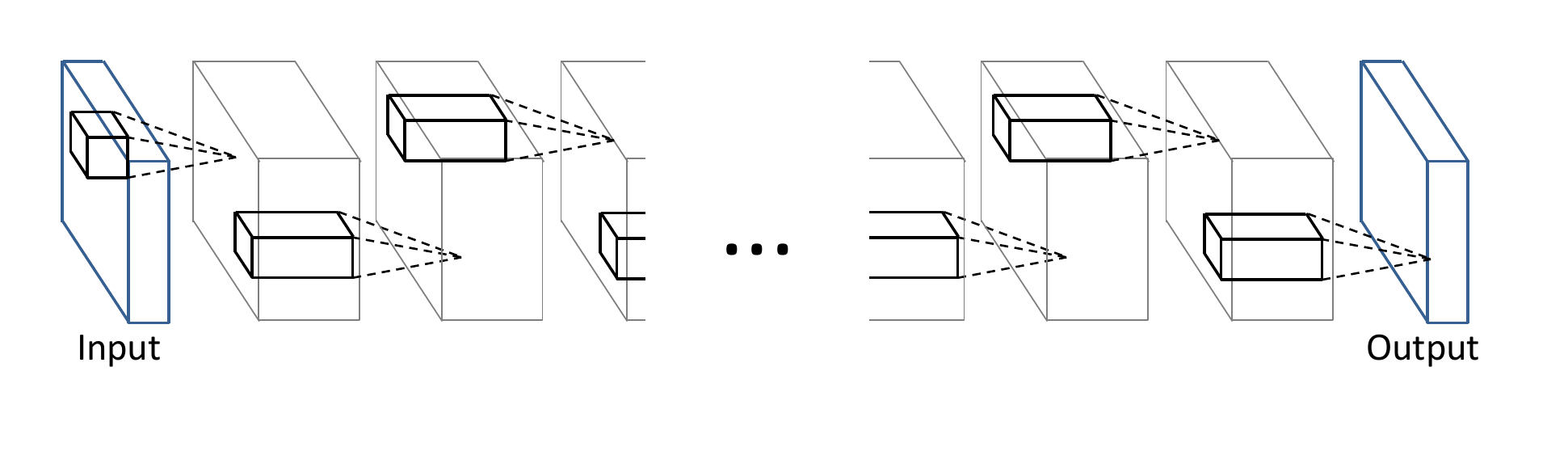}
\label{fig:fcn}
}
\end{tabular}
\caption{Key components of the proposed network.}
\label{fig:intro}
\end{center}
\end{figure}

Lately, there have been increasing efforts to use deep learning based approaches for hyperspectral image (HSI) classification~\cite{YChenJSTAR14,WHuJS15,WZhaoTGARS16,YChenTGARS16,PLiuJSTAR17,PZhongJSTAR17,YChenJSTAR15,TLiICIP14}.  However, in reality, large scale HSI datasets are not currently commonly available, which leads to sub-optimal learning of DCNN with large numbers of parameters due to the lack of enough training samples.  The limited access to large scale hyperspectral data has been preventing existing CNN-based approaches for HSI classification~\cite{YChenJSTAR14,WZhaoTGARS16,WHuJS15,YChenTGARS16,PLiuJSTAR17,PZhongJSTAR17} from leveraging {\it deeper} and {\it wider} networks that can potentially better exploit very rich spectral and spatial information contained in hypersepctral images. Therefore, current state-of-the-art CNN-based approaches mostly focus on using small-scale networks with relatively fewer numbers of layers and nodes in each layer at the expense of a decrease in performance.  Deeper and wider mean using relatively larger numbers of layers (depth) and nodes in each layer (width), respectively.  Accordingly, the reduction of the spectral dimension of the hyperspectral images is in general initially performed to fit the input data into the small-scale networks by using techniques, such as principal component analysis (PCA)~\cite{KPearsonPM1901}, balanced local discriminant embedding (BLDE)~\cite{WZhaoTGARS16}, pairwise constraint discriminant analysis and nonnegative sparse divergence (PCDA-NSD)~\cite{XWangJSTAR17}, etc.  However, leveraging large-scale networks is still desirable to jointly exploit underlying nonlinear spectral and spatial structures of hyperspectral data residing in a high dimensional feature space.  In the proposed work, we aim to build a deeper and wider network given limited amounts of hypersectral data that can jointly exploit spectral and spatial information together. To tackle issues associated with training a large scale network on limited amounts of data, we leverage a recently introduced concept of ``residual learning'', which has demonstrated the ability to significantly enhance the train efficiency of large scale networks.  The residual learning~\cite{KHeCVPR16} basically reformulates the learning of subgroups of layers called modules in such a way that each module is optimized by the residual signal, which is the difference between the desired output and the module input, as shown in Figure~\ref{fig:residual_learning}.  It is shown that the residual structure of the networks allows for considerable increase in depth and width of the network leading to enhanced learning and eventually improved generation performance. Therefore, the proposed network does not require pre-processing of dimensionality reduction of the input data as opposed to the current state-of-the art techiniques.

To achieve the state-of-the art performance for HSI classification, it is essential that spectral and spatial features are jointly exploited.  As can be seen in \cite{YChenJSTAR14,WZhaoTGARS16,WHuJS15,YChenJSTAR15,TLiICIP14}, the current state-of-the-art approaches for deep learning based HSI classification fall short of fully exploiting spectral and spatial information together.  The two different types of information, spectral and spatial, are more or less acquired separately from pre-processing and then processed together for feature extraction and classification in~\cite{YChenJSTAR14,YChenJSTAR15}.  Hu et al.~\cite{WHuJS15} also failed to jointly process the spectral and spatial information by only using individual spectral pixel vectors as input to the CNN.  In this paper, inspired by~\cite{CSzegedyCVPR15}, we propose a novel deep learning based approach that uses fully convolutional layers (FCN)~\cite{JLongCVPR15} to better exploit spectral and spatial information from hyperspectral data.  At the initial stage of the proposed deep CNN, a multi-scale convolutional filter bank conceptually similar to the ``inception module'' in~\cite{CSzegedyCVPR15} is simultaneously scanned through local regions of hyperspectral images generating initial spatial and spectral feature maps.  The multi-scale filter bank is basically used to exploit various local spatial structures as well as local spectral correlations.  The initial spatial and spectral feature maps generated by applying the filter bank are then combined together to form a joint spatio-spectral feature map, which contains rich spatio-spectral characteristics of hyperspectral pixel vectors.  The joint feature map is in turn used as input to subsequent layers that finally predict the labels of the corresponding hyperspectral pixel vectors.

The proposed network\footnote{A preliminary version of this paper \cite{HLeeIGARSS16} was presented at the 2016 IEEE International Geoscience and Remote Sensing Symposium (IGARSS 2016).} is an end-to-end network, which is optimized and tested all together without additional pre- and post-processing.  The proposed network is a fully convolutional network (FCN)~\cite{JLongCVPR15} (Figure~\ref{fig:fcn}) to take input hyperspectral images of arbitrary size and does not use any subsampling (pooling) layers that would otherwise result in the output with different size than the input; this means that the network can process hyperspectral images with arbitrary sizes.  In this work, we evaluate the proposed network on three benchmark datasets with different sizes (145$\times$145 pixels for the Indian Pines dataset, 610$\times$340 pixels for the University of Pavia dataset, and 512$\times$217 for the Salinas dataset).  The proposed network is composed of three key components; a novel fully convolutional network, a multi-scale filter bank, and residual learning as illustrated in Figure~\ref{fig:intro}.  Performance comparison shows enhanced classification performance of the proposed network over the current state-of-the-art on the three datasets.

The main contributions of this paper are as follows:
\begin{itemize}
\item We introduce the {\it deeper} and {\it wider} network with the help of ``residual learning'' to overcome sub-optimality in network performance caused primarily by limited amounts of  training samples.
\item We present a novel deep CNN architecture that can jointly optimize the spectral and spatial information of hyperspectral images.
\item The proposed work is one of the first attempts to successfully use a very deep fully convolutional neural network for hyperspectral classification.
\end{itemize}

The remainder of this paper is organized as follows.  In Section~\ref{sec:rel_works}, related works are described. Details of the proposed network are explained in Section~\ref{sec:proposed_network}.  Performance comparisons among the proposed network and current sate-of-the-art approaches are described in Section~\ref{sec:exp}.   The paper is concluded in Section~\ref{sec:concl}.

\section{Related Works}
\label{sec:rel_works}

\subsection{Going deeper with Deep CNN for object detection/classification}

LeCun, et al. introduced the first deep CNN called LeNet-5~\cite{YLeCunNC1989} consisting of two convolutional layers, two fully connected layers, and one Gaussian connection layer with additional several layers for pooling.  With the recent advent of large scale image databases and advanced computational technology, relatively deeper and wider networks, such as AlexNet~\cite{AKrizhevskyNIPS12}, began to be constructed on large scale image datasets, such as ImageNet~\cite{JDengCVPR09}.  AlexNet used five convolutional layers with three subsequent fully connected layers.  Simonyan and Zisserman~\cite{KSimonyanICLR15} significantly increased the depth of Deep CNN, called VGG-16, with 16 convolutional layers.  Szegedy et al.~\cite{CSzegedyCVPR15} introduced a 22 layer deep network called GoogLeNet, by using multi-scale processing, which is realized by using a concept of ``inception module.''  He et al.~\cite{KHeCVPR16} built a network substantially deeper than those used previously by using a novel learning approach called ``residual learning'', which can significantly improve training efficiency of deep networks.

\begin{figure*}[t]
\begin{center}
\includegraphics[trim = 5mm 5mm 5mm 5mm,width=\linewidth]{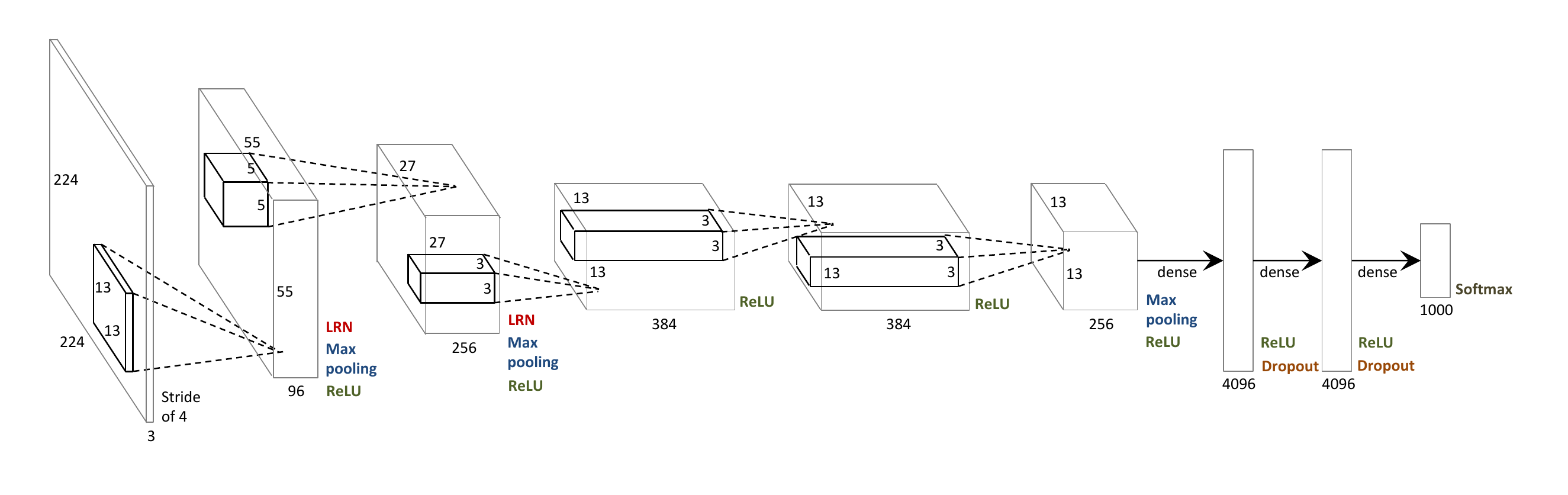}
\caption{{\bf AlexNet~\cite{AKrizhevskyNIPS12}.} The network consists of five convolutional layers and three fully connected layers.  In the illustration, cubes and boxes indicate data blobs.  Several non-linear functions are also used in the network.  Non-linear functions are listed beside the output blobs of each layer in order.}
\label{fig:alexnet}
\end{center}
\end{figure*}

\subsection{Deep CNN for Hyperspectral Image Classification}

A large number of approaches have been developed to tackle HSI classification problems~\cite{PGurramTGARS13,YGuTGARS16,FMorsierTGARS16,JLiuTGARS16,QWangTGARS16,BGuoTIP08,LYangJSTAR17,RRoscherTGARS16,JinlinLiuTGARS16,AZhetabianTGARS16,SJiaTGARS16,JunshiXiaTGARS16,ZZhongTGARS16,JXiaTGARS16,HYangTGARS16,MToksozTGARS16,PZhongTIP10,KBernardTIP12,YGaoTIP14,YChenTGARS16,MBrellTGARS17,SJiaTGARS17a,SJiaTGARS17b,SMeiJSTAR17,HSuJSTAR17}.    Recently, kernel methods, such as multiple kernel learning~\cite{PGurramTGARS13,YGuTGARS16,FMorsierTGARS16,JLiuTGARS16,QWangTGARS16,BGuoTIP08,LYangJSTAR17}, have been widely used primarily because they can enable a classifier to learn a complex decision boundary with only a few parameters.  This boundary is built by projecting the data onto a high-dimensional reproducing kernel Hilbert space~\cite{EStroblICMLA13}.  This makes it suitable for exploiting dataset with limited training samples.  However, recent advance of deep learning-based approaches has shown drastic performance improvements because of its capabilities that can exploit complex local nonlinear structures of images using many layers of convolutional filters.  To date, several deep learning-based approaches~\cite{YChenJSTAR14,WZhaoTGARS16,WHuJS15,YChenTGARS16,PLiuJSTAR17,PZhongJSTAR17} have been developed for HSI classification.  But few have achieved breakthrough performance due mainly to sub-optimal learning caused by the lack of enough training samples and the use of relatively small scale networks.  

Deep learning approaches normally require large scale datasets whose size should be proportional to the number of parameters used by the network to avoid overfitting in learning the network.  Chen et al.~\cite{YChenJSTAR14} used stacked autoencoders (SAE) to learn deep features of hyperspectral signatures in an unsupervised fashion followed by logistic regression used to classify extracted deep features into their appropriate material categories.  Both a representative spectral pixel vector and the corresponding spatial vector obtained from applying principle component analysis (PCA) to hyperspectral data over the spectral dimension are acquired separately from a local region and then jointly used as an input to the SAE.  In~\cite{YChenJSTAR15}, Chen et al. replaced SAE by a deep belief network (DBN), which is similar to the deep convolutional neural network for HSI classification.  Li et al.~\cite{TLiICIP14} also used a two-layer DBN but did not use initial dimensionality reduction, which would inevitably cause the loss of critical information of hyperspectral images.  Hu et al.~\cite{WHuJS15} fed individual spectral pixel vectors independently through simple CNN, in which local convolutional filters are applied to the spectral vectors extracting local spectral features.  Convolutional feature maps generated after max pooling are then used as the input to the fully connected classification stage for material classification.  Chen et al.~\cite{YChenTGARS16} also used deep convolutional neural network adopting five convolutional layers and one fully connected layer for hyperspectral classification.  

Unlike these deep learning-based approaches, we first attempt to build much deeper and wider network using relatively small amounts of training samples.  Once the network is effectively optimized, it is expected to provide enhanced performance over relatively shallow and narrow networks.

\section{The Contextual Deep Convolutional Neural Network}
\label{sec:proposed_network}

In this section, we first describe the widely used CNN model referred to as AlexNet and then discuss the overall architecture of the proposed network. We elaborate on the two key components of the proposed network, ``multi-scale convolutional filter bank'' and ``residual learning.''  The learning process of the network is discussed at the end of the section.

\begin{figure*}[t]
\begin{center}
\includegraphics[trim = 5mm 5mm 5mm 5mm,width=\linewidth]{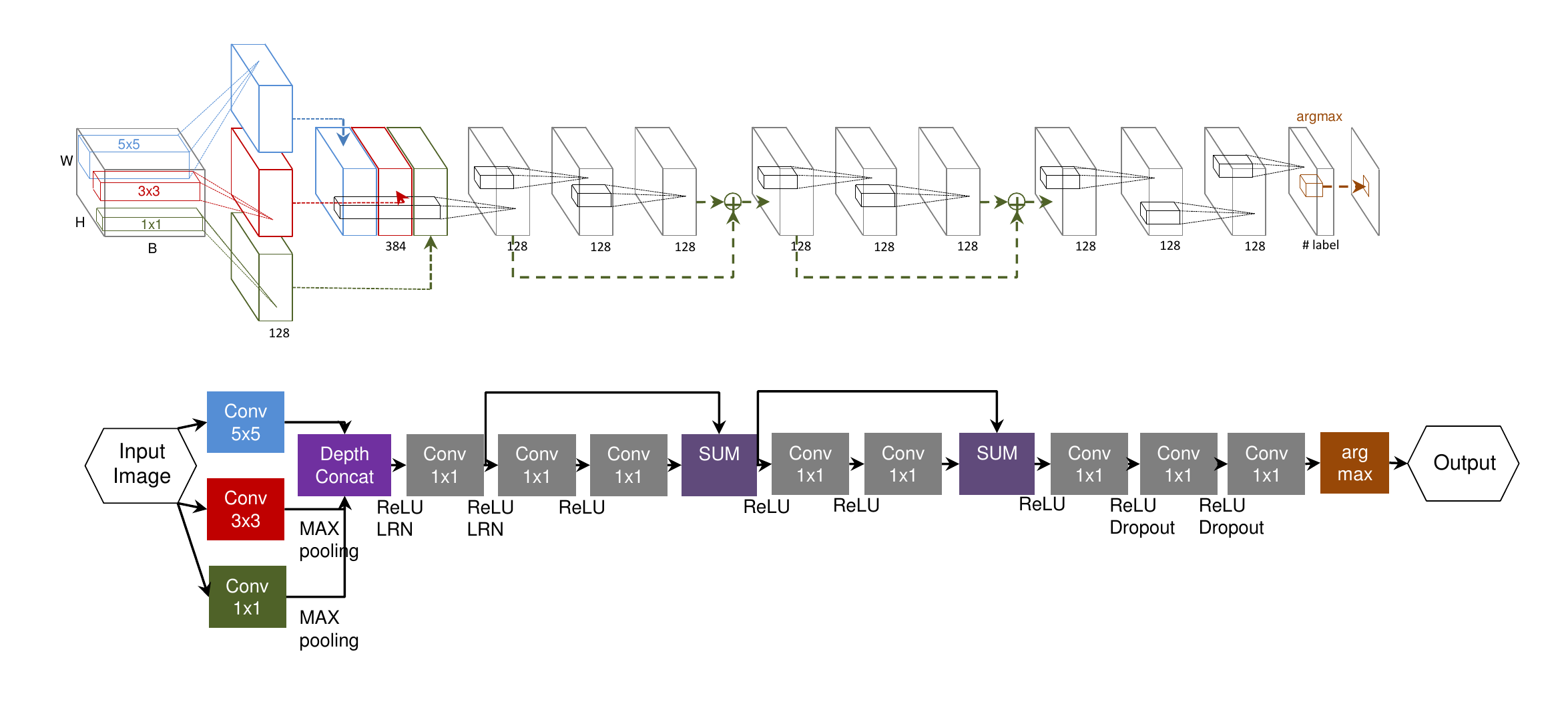}
\caption{{\bf An illustration of the architecture of the proposed network.} The first row illustrates input and output blobs of convolutional layers and their connections.  The number of filters of each convolutional layer is indicated under its output blob.  The second row shows a flow chart of the network.}
\label{fig:architecture}
\end{center}
\end{figure*}

\subsection{Deep Convolutional Neural Network}
\label{ssec:cnn}

A widely used deep CNN model includes multiple layers of neurons, each of which extracts a different level of non-linear features from the input ranging from low to high level features.  Non-linearity in each layer is achieved by applying a nonlinear activation function to the output of local convoultional filters in each layer.  The proposed network is basically a convolutional neural network with a nonlinear activation function used in~\cite{AKrizhevskyNIPS12}.

In this section, we first describe the architecture of AlexNet, a widely used deep CNN model, as shown in Figure~\ref{fig:alexnet}, to provide the basis for understanding the architecture of the proposed network. AlexNet consists of five convolutional layers and three fully connected layers.  Each fully connected layer contains linear weights $W_{FC}$ connecting the relationship between input $x$ and output $y$:

\begin{equation}
y=W_{FC}\cdot x,
\end{equation}
where $x$ and $y$ represent the input and output vectors.  A convolutional layer with $N$ local filters, $W_{C,i}, i=1,2,...,N$, extracts local nonlinear features from the input and is expressed as:

\begin{equation}
y=\{W_{C,i}\ast x\}_{i=1,2,...,N},
\end{equation}
where $\ast$ denotes a convolution.  The filter size of all $\{W_{C,i}\}_{i=1,2,...,N}$ is carefully determined to be much smaller than the size of $W_{FC}$.

In~\cite{AKrizhevskyNIPS12}, several non-linear components, such as the {\it local response normalization (LRN)}, {\it max pooling}, the {\it rectified linear unit (ReLU)}, {\it dropout}, and {\it softmax} are used.  {\it LRN} normalizes each activation $a_i$ over local activations of $n$ adjacent filters centered on the position $(p_x,p_y)$, which aims to generalize filter responses, 

\begin{equation}
a_i^{*}(p_x,p_y)=a_i(p_x,p_y)/\bigg(k+\alpha\sum_{j=i-n/2}^{i+n/2}{\Big(a_j(p_x,p_y)\Big)^2}\bigg)^\beta,
\end{equation}
where $k$, $n$, $\alpha$, and $\beta$ are hyper-parameters.  {\it Max pooling} down-samples the output of layers by replacing a sub-region of the output with the maximum value, which is commonly used for dimensionality reduction in CNN.  {\it ReLU} rectifies negative values to zero and is used for the network to learn parameters with positive activations only.  {\it ReLU} basically replaces the sigmoid function commonly used for other neural networks mainly because learning deep CNN with {\it ReLU} is several times faster than the network with other nonlinear activation functions such as $tanh$.  {\it Dropout} is a function that forces the output of individual nodes of each layer to be zero with a
probability under a certain threshold, which takes any value within (0, 1).  In this work, we used a threshold of 0.5.  {\it Dropout} reduces overfitting by preventing multiple adaptations of training data simultaneously (referred to as ``complex co-adaptions'').  {\it Softmax} is a generalization of the logistic function, which is defined as the gradient-log-normalizer of the categorical probability distribution:

\begin{equation}
P(y=j | x, \{f_{k}\}_{k=1,2,\cdots,K}) = \frac{e^{f_{j}(x)}}{\sum_{k=1}^{K}{e^{f_{k}(x)}}},
\end{equation}
where $f_{j}$ is a classification function for a $j^{th}$ class, whose input and output are $x$ and $y$, respectively.  Therefore, {\it softmax} is useful for probabilistic multiclass classification including HSI classification.

\subsection{Architecture of the Proposed Network}
\label{ssec:architecture}

We propose a novel fully convolutional network (FCN)~\cite{JLongCVPR15} with a number of convolutional layers for HSI classification, as show in Figure~\ref{fig:architecture}.  The first part of the network is a ``multi-scale filter bank'' followed by two blocks of convolutional layers associated with residual learning.  The last three convolutional layers function in a similar manner to the fully conected layers for classification of the AlexNet, which performs classification using local features.  Similar to AlexNet, the $7^{th}$ and $8^{th}$ convolutional layers have {\it dropout} in training.  The {\it ReLU} is used after the multi-scale filter bank, the $2^{th}$, $3^{rd}$, $5^{th}$, $7^{th}$, $8^{th}$ convolutional layers, and two residual learning modules.  The output of the first two convolutional layers is normalized by {\it LRN}.  Note that the height and width of all data blobs in the architecture are the same and only their depth changes.  No dimensionality reduction is performed throughout the FCN processing.  

\begin{figure}[t]
\begin{center}
\includegraphics[trim = 5mm 5mm 5mm 5mm,width=\linewidth]{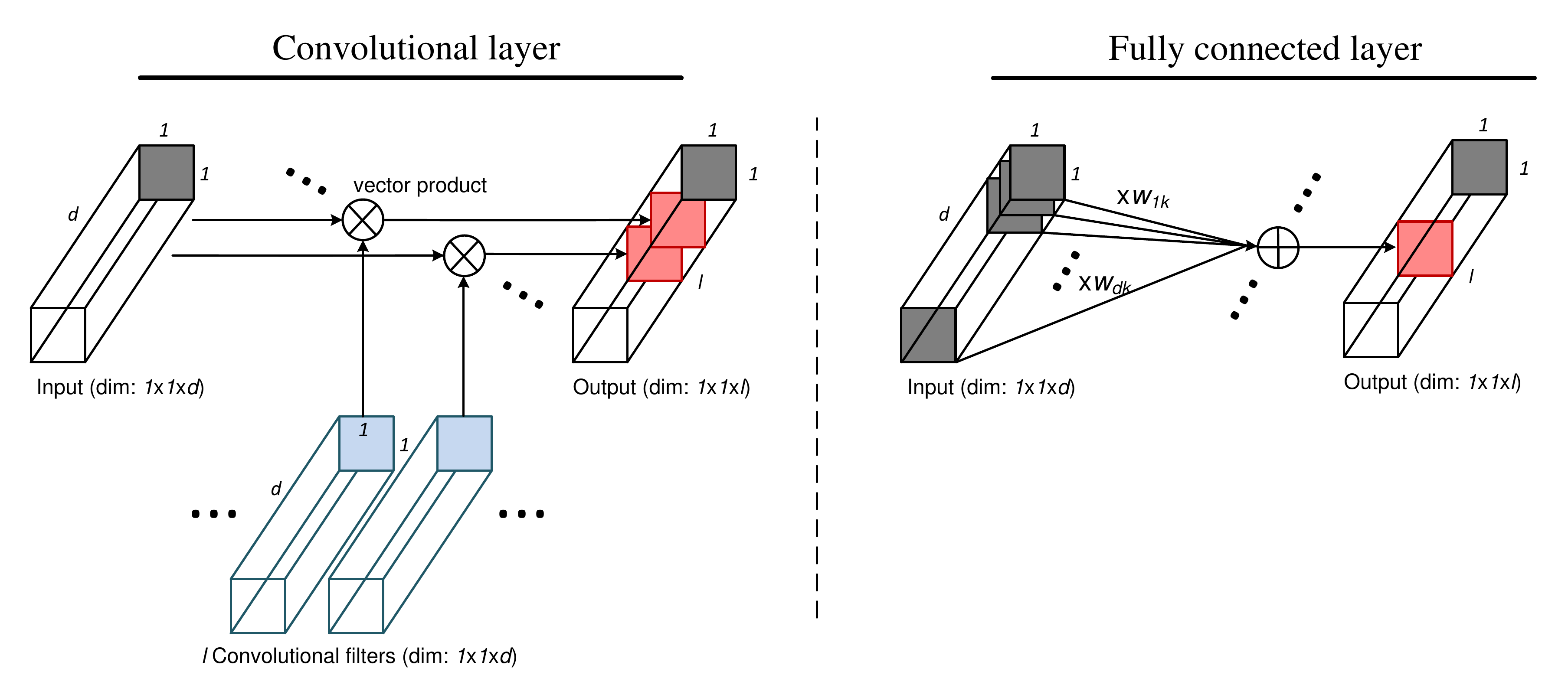}
\caption{{\bf Convolutionalized model.} For pixel classification, a convolutional layer can achieve the same effect as the fully connected layer with the same number of weights. In the above illustration, the convolutional layer uses $l$ convolutional filters whose dimension is $1\times 1\times d$ and weights of the fully connected layer is $\{w_{i,j}\}_{i=1,\cdots,d, j=1,\cdots,l}$.  Both convolutional layer and fully connected layer use $d\times l$ weights.}
\label{fig:convolutionalization}
\end{center}
\end{figure}

Note that convolving a $1\times 1\times d$ blob with $l$ filters whose size is $1\times 1\times d$ can achieve the same effect as fully connecting the $1\times 1\times d$ input blob to $l$ output nodes, as illustrated in Figure~\ref{fig:convolutionalization}.  Due to this ``convolutionalized model'', FCN can be used for pixel classification, such as semantic segmentation, HSI classification, etc.  Since our network is based on FCN, the proposed network learns on $5\times 5$ pixels centered on individual pixel vectors and is applied to the whole image in test.\\

\noindent {\bf How Much Deeper Does the Proposed Network Go?} The proposed network contains a total of 9 layers, which is much deeper than other CNNs for HSI classification trained on the same datasets \cite{WHuJS15}.  However, the depth of 9 still does not seem to be large enough, especially when compared to the current state-of-the-art CNNs for image classification, such as ResNet \cite{KHeCVPR16}.  This is mainly because HSI-based CNNs have to be trained on much smaller amounts of training samples than that of the image classification CNNs primarily trained on large scale databases, such as ImageNet (1.2 M) \cite{JDengCVPR09}.  Constrained by highly limited HSI training data, the proposed {\it going deeper} strategy opts not to use a very large number of layers to avoid overfitting.  However, it still uses a much greater number of layers than that of any other HSI-based CNNs.  Table \ref{tab:depth_comparison} shows a comparison of various CNNs for both image and HSI classification with regards to network variables, such as the number of layers and parameters, training data size, and a ratio between the number of the parameters and data size. 

\begin{table}[t]
\caption{Comparison of network variables of various CNNs for both image and HSI classification.}
\begin{center}
\rowcolors{0}{}{lightgray}
\setlength{\tabcolsep}{3.5pt}
\renewcommand{\arraystretch}{1.4}
\begin{tabular}{lrrrr}
\hline
Method & \# of Layer & param & data size & param/data \\
\hline\hline
AlexNet~\cite{AKrizhevskyNIPS12} & 8 & 59.3M & 12M & 4.94 \\
VGG16~\cite{KSimonyanICLR15} & 16 & 135.1M & 12M & 11.26 \\
GoogLeNet~\cite{CSzegedyCVPR15} & 22 & 6.8M & 12M & 0.57 \\
ResNet152~\cite{KHeCVPR16} & 152 & 56.0M & 12M & 4.66 \\\hline
\cite{WHuJS15}-Indian Pines & 3 & 79.5K & 1.6K & 49.69 \\
\cite{WHuJS15}-Salinas & 3 & 80.3K & 3.1K & 25.90 \\
\cite{WHuJS15}-U. of Pavia & 3 & 59.8K & 1.8K & 33.22 \\
The Proposed-Indian Pines & 9 & 1122.5K & 6.4K & 175.39 \\
The Proposed-Salinas & 9 & 1875.8K & 12.4K & 151.27 \\
The Proposed-U. of Pavia & 9 & 610.6K & 7.2K & 84.81 \\
\hline
\end{tabular}
\end{center}
\label{tab:depth_comparison}
\end{table}

Similar to data augmentation used in image classification CNNs, the proposed network also uses a data augmentation strategy described in Section \ref{ssec:learning}. As shown Table \ref{tab:depth_comparison}, the proposed network provides much larger ratios between the number of parameters and training data size than those of the baseline \cite{WHuJS15} for the same training dataset. Also, the parameter vs. data ratios of the proposed networks are at least approximately eight times larger than that of any image classification CNNs. This indicates that the architecture of the proposed network is designed to ensure that it provides sufficient depth of layers to fully exploit training data. 

\begin{figure*}[t]
\begin{center}
\includegraphics[trim = 10mm 5mm 5mm 5mm,width=\linewidth]{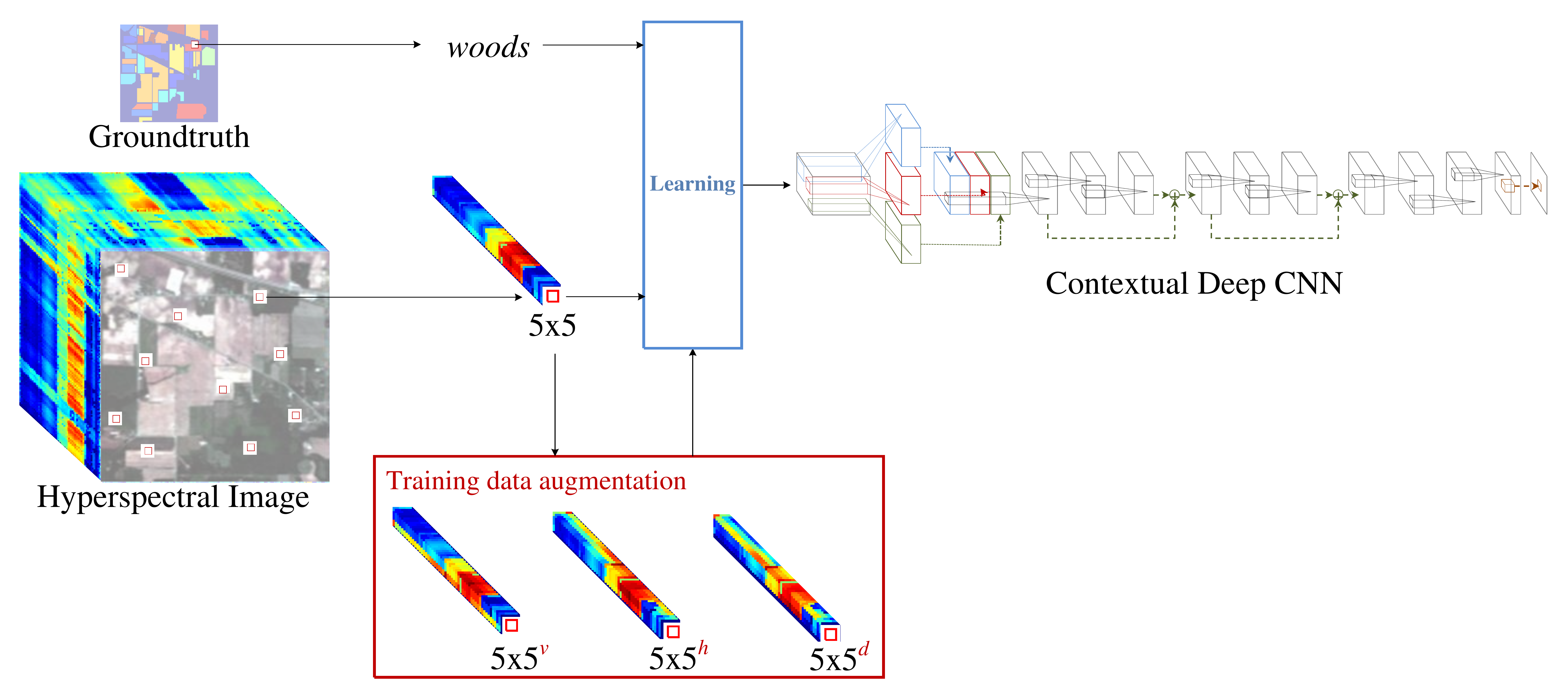}
\caption{{\bf The learning process of the proposed network.} In the hyperspectral image, 1$\times$1 training pixel and its neighboring 5$\times$5 pixels are indicated by a red and white rectangle, respectively.  In the red box representing augmented training data, 5$\times$5$^{{\color{red} v}}$, 5$\times$5$^{{\color{red} h}}$, and 5$\times$5$^{{\color{red} d}}$ are the training samples mirrored across across the horizontal, vertical, and diagonal axes, respectively.}
\label{fig:learn}
\end{center}
\end{figure*}

\subsection{Multi-scale Filter Bank}
\label{ssec:inception}

The first convolutional layer applied to the input hyperspectral image uses a multi-scale filter bank that locally convolves the input image with three convolutional filters with different sizes ($1\times 1 \times B$, $3\times 3 \times B$, and $5\times 5\times B$ where $B$ is the number of spectral bands).  The $3\times 3 \times B$ and $5\times 5\times B$ filters are used to exploit local spatial correlations of the input image while the $1\times 1 \times B$ filters are used to address spectral correlations.  The output of the first convolutional layer, the three convolutional feature maps, as shown in Figure~\ref{fig:architecture}, are combined together to form a joint spatio-spectral feature map used as input to the subsequent convolutional layers.  

However, since the size of the feature maps from the three convolutional filters is different from each other, a strategy to adjust the size of the feature maps to be same to combine them into a joint feature map is needed.  First, a space of two-pixel width filled with zeros is padded around the input image such that the size of the feature maps from the $1\times 1$, $3\times 3$, and $5\times 5$ filters becomes $(H+4,W+4)$, $(H+2,W+2)$, and $(H,W)$, respectively.  $H$ and $W$ are the height and width of the input image, respectively.  The size of all the feature maps becomes  $(H,W)$ after $5\times 5$ and $3\times 3$ max poolings are applied to the feature maps from the $1\times 1$ and $3\times 3$ filters, respectively.

$3\times 3$ and $5\times 5$ convolutions with a large number of spectral bands can be expensive and merging of the output of the convolutional filter bank causes the size of the network to increase, which also inevitably leads to high computational complexity.  As the network size is increased, optimizing the network with a small number of training samples will face overfitting and divergence.  Therefore, a strategy to address the above issues needs to be used.  To tackle the issues, we use training data augmentation and residual learning modules described in  Section~\ref{ssec:res_learning} and~\ref{ssec:learning}.\\

\noindent {\bf Functionality of the Multi-scale Filter Bank.} The multi-scale filter bank conceptually similar to the inception module in \cite{CSzegedyCVPR15} is used to optimally exploit diverse local structures of the input image.  \cite{CSzegedyCVPR15} demonstrates the effectiveness of the inception module that enables the network to get deeper as well as to exploit local structures of the input image achieving state-of-the-art performance in image classification.  The multi-scale filter bank in the proposed network is used in a somewhat different manner that aims to jointly exploit local spatial structures in conjunction with local spectral correlations at the initial stage of the proposed structure.

\subsection{Residual Learning}
\label{ssec:res_learning}

The subsequent convolutional layers use $1\times 1 \times B$ filters to extract nonlinear features from the joint spatio-spectral feature map.  We use two modules of ``residual learning''~\cite{KHeCVPR16}, which is shown to help significantly improve training efficiency of deep networks.  The residual learning is to learn layers with reference to the layer input using the following formula:

\begin{equation}
y=\mathcal{F}(x,\{W_i\})+x,
\end{equation}
where $x$ and $y$ are the input and output vectors of the layers considered, respectively.  The function $\mathcal{F}:=y-x$ is the residual mapping of the input to the residual output $y-x$ using convolutional filters $W_i$.  \cite{KHeCVPR16} proved that it is easier to optimize $W_i$ with the residual mapping than to optimize those weights with the unreferenced mapping.  In the proposed network, two convolutional layers are used for the residual mapping, which is called ``shortcut connections''.  The residual learning is very effective in practice, which is also proven in~\cite{KHeCVPR16}.  {\it ReLU} is the function that makes the first layer in the module nonlinear.  Note that both the multi-scale filter bank and the residual learning are effective in increasing the depth and width of the network while keeping the computational budget constrained~\cite{CSzegedyCVPR15,KHeCVPR16}.  This helps to effectively learn the deep network with a small number of training samples.

\subsection{Learning the Proposed Network}
\label{ssec:learning}

We randomly sample a certain number of pixels from the hyperspectral image for training and use the rest to evaluate the performance of the proposed network.  For each training pixel, we crop surrounding 5$\times$5 neighboring pixels for learning convolutional layers.  The proposed network contains approximately 1000K parameters, which are learned from several hundreds of training pixels from each material category.  To avoid overfitting, we augment the number of training samples four times by mirroring the training samples across the horizontal, vertical, and diagonal axes.  Figure~\ref{fig:learn} illustrates the learning process of the proposed network.

For learning the proposed network, stochastic gradient descent (SGD) with a batch size of 10 samples is used with 100K iterations, a momentum of 0.9, a weight decay of 0.0005 and a gamma of 0.1.  We initially set a base learning rate as 0.001.  The base learning rate is decreased to 0.0001 after 33,333 iterations and is further reduced to 0.00001 after 66,666 iterations.  To learn the network, the last argmax layer is replaced by a softmax layer commonly used for learning convolutional layers.  The first, second, and ninth convolutional layers are initialized from a zero-mean Gaussian distribution with standard deviation of 0.01 and the remaining convolutional layers are initialized with standard deviation of 0.005.  Biases of all convolutional layers except the last layer are initialized to one and the last layer is initialized to zero.

\section{Experimental Results}
\label{sec:exp}

\subsection{Dataset and Baselines}
\label{ssec:dataset}

\begin{figure}[t]
\begin{center}
\includegraphics[trim = 6mm 10mm 7mm 5mm,width=\linewidth]{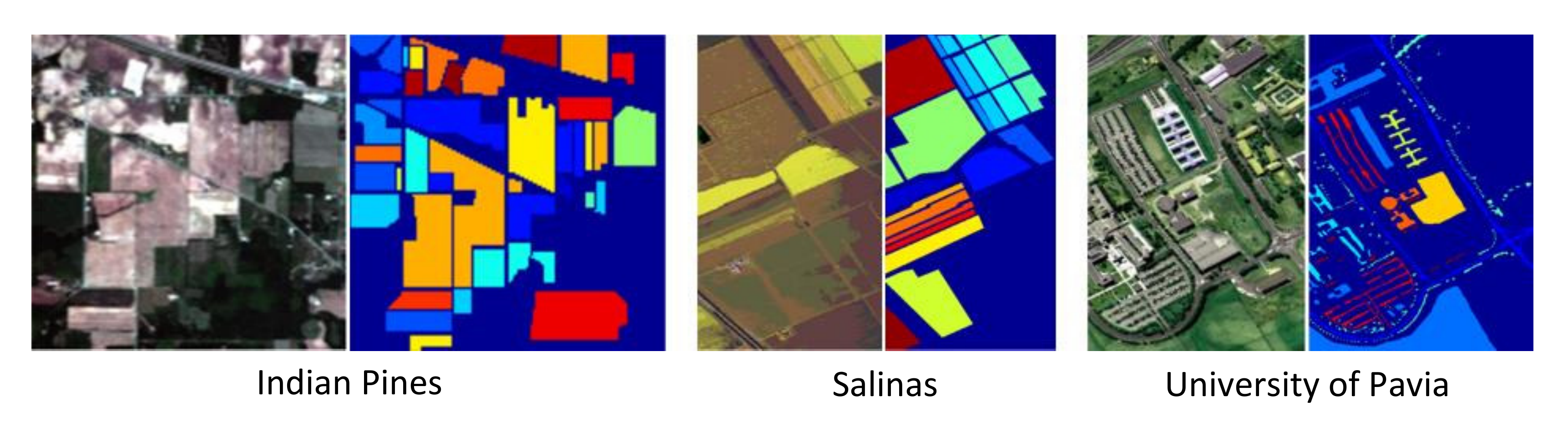}
\caption{{\bf Three HSI datasets.} Indian pines, Salinas, and University of Pavia datasets.  For each dataset, three-band color composite image is given on the left and ground truth is shown on the right.  In groundtruth, pixels belonged to the same class are depicted with the same color.}
\label{fig:hyp_img}
\end{center}
\end{figure}

\begin{figure*}[t]
\begin{center}
\begin{tabular}{ccc}
\subfloat[Indian Pines]{
\includegraphics[trim = 17mm 6mm 15mm 5mm,width=.395\textwidth]{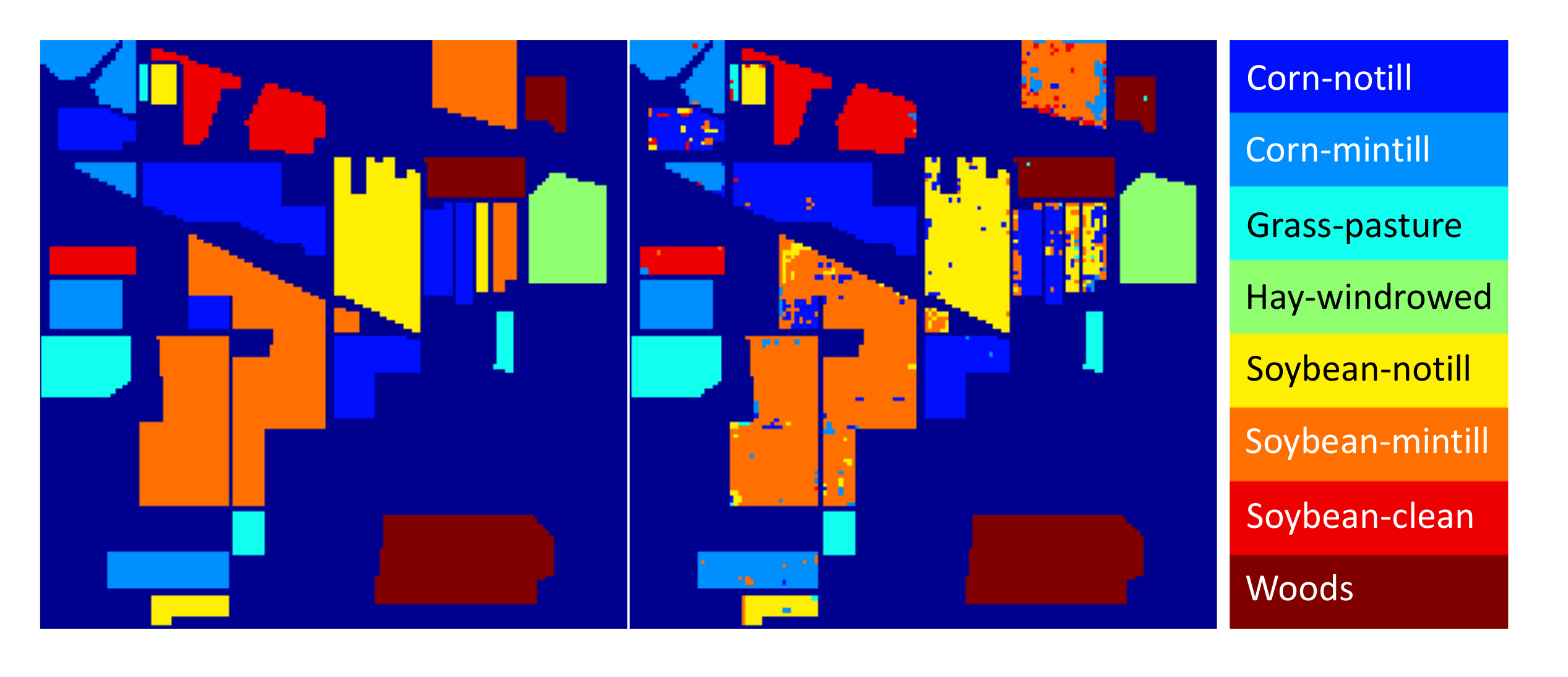}
\label{fig:qual_indian_pines}} &
\subfloat[Salinas]{
\includegraphics[trim = 17mm 5mm 10mm 5mm,width=.265\textwidth]{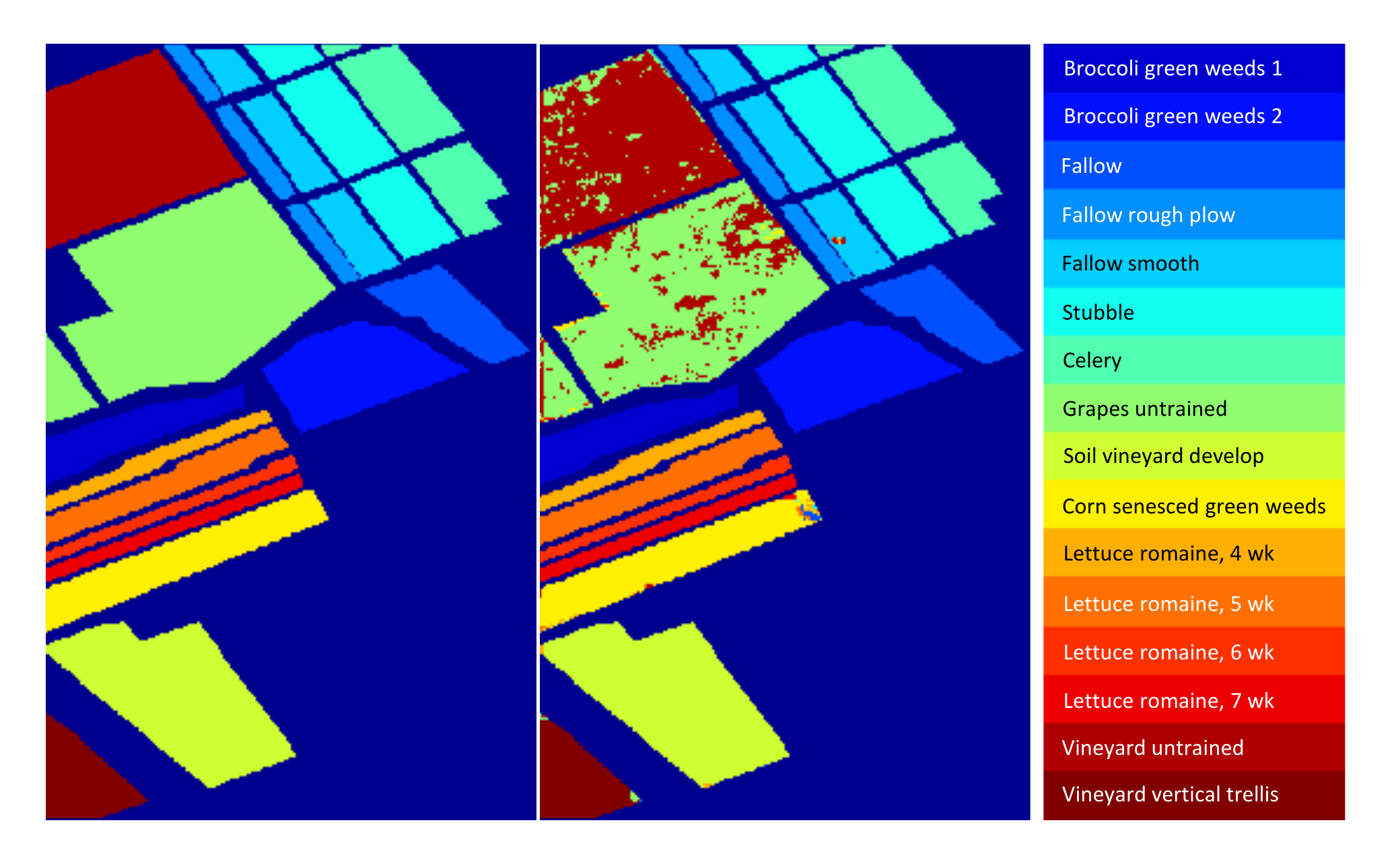}
\label{fig:qual_salinas}} &
\subfloat[University of Pavia]{
\includegraphics[trim = 17mm 9mm 15mm 5mm,width=.222\textwidth]{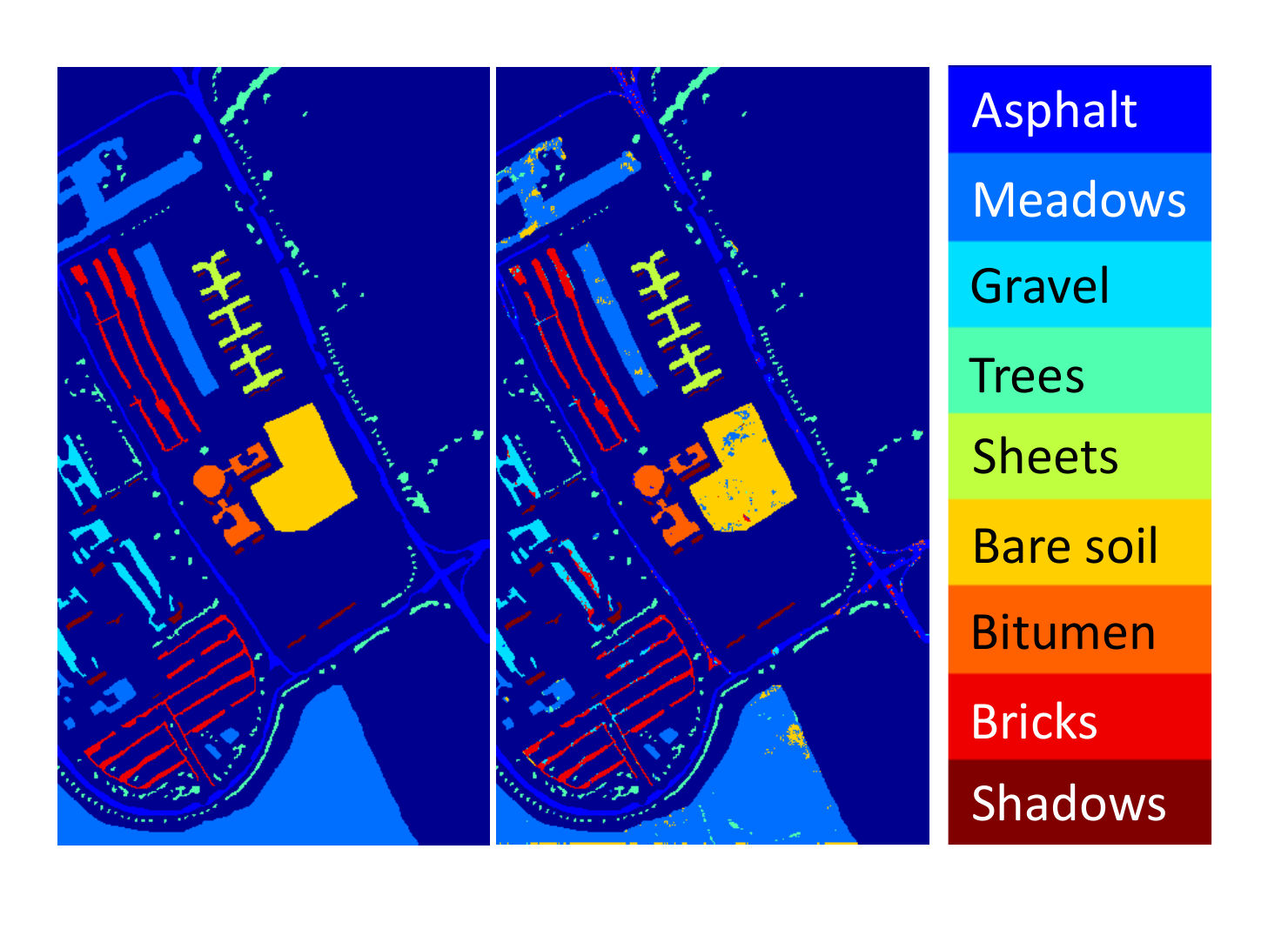}
\label{fig:qual_upavia}}
\end{tabular}
\caption{RGB composition maps of groundtruth (left) of each dataset and the classification results (center) from the proposed network for the dataset.}
\label{fig:qual}
\end{center}
\end{figure*}

\begin{table}[t]
\caption{Selected classes for evaluation and the numbers of training and test samples used from the Indian Pines dataset}
\vspace{-0.5cm}
\begin{center}
\rowcolors{0}{}{lightgray}
\setlength{\tabcolsep}{18.5pt}
\renewcommand{\arraystretch}{1.4}
\begin{tabular}{llcc}
\hline
No & Class & Training & Test \\
\hline
1 & Corn-notill & 200 & 1228 \\
2 & Corn-mintill & 200 & 630 \\
3 & Grass-pasture & 200 & 283 \\
4 & Hay-windrowed & 200 & 278 \\
5 & Soybean-notill & 200 & 772 \\
6 & Soybean-mintill & 200 & 2255 \\
7 & Soybean-clean & 200 & 393 \\
8 & Woods & 200 & 1065 \\
\hline
& Total & 1600 & 6904 \\
\hline
\end{tabular}
\end{center}
\label{tab:split_indian_pine}
\end{table}

\begin{table}[t]
\caption{Selected classes for evaluation and the numbers of training and test samples used from the Salinas dataset}
\begin{center}
\rowcolors{0}{}{lightgray}
\setlength{\tabcolsep}{13pt}
\renewcommand{\arraystretch}{1.4}
\begin{tabular}{llcc}
\hline
No & Class & Training & Test \\
\hline
1 & Broccoli green weeds 1 & 200 & 1809 \\
2 & Broccoli green weeds 2 & 200 & 3526 \\
3 & Fallow & 200 & 1776 \\
4 & Fallow rough plow & 200 & 1194 \\
5 & Fallow smooth & 200 & 2478 \\
6 & Stubble & 200 & 3759 \\
7 & Celery & 200 & 3379 \\
8 & Grapes untrained & 200 & 11071 \\
9 & Soil vineyard develop & 200 & 6003 \\
10 & Corn senesced green weeds & 200 & 3078 \\
11 & Lettuce romaines, 4 wk & 200 & 868 \\
12 & Lettuce romaines, 5 wk & 200 & 1727 \\
13 & Lettuce romaines, 6 wk & 200 & 716 \\
14 & Lettuce romaines, 7 wk & 200 & 870 \\
15 & Vineyard untrained & 200 & 7068 \\
16 & Vineyard vertical trellis & 200 & 1607 \\
\hline
& Total & 3200 & 50929 \\
\hline
\end{tabular}
\end{center}
\label{tab:split_salinas}
\end{table}

\begin{table}[t]
\caption{Selected classes for evaluation and the numbers of training and test samples used from the University of Pavia dataset}
\vspace{-0.55cm}
\begin{center}
\rowcolors{0}{}{lightgray}
\setlength{\tabcolsep}{20.5pt}
\renewcommand{\arraystretch}{1.4}
\begin{tabular}{llcc}
\hline
No & Class & Training & Test \\
\hline
1 & Asphalt & 200 & 6431 \\
2 & Meadows & 200 & 18449 \\
3 & Gravel & 200 & 1899 \\
4 & Trees & 200 & 2864 \\
5 & Sheets & 200 & 1145 \\
6 & Bare soils & 200 & 4829 \\
7 & Bitumen & 200 & 1130 \\
8 & Bricks & 200 & 2482 \\
9 & Shadows & 200 & 747 \\
\hline
& Total & 1800 & 40976 \\
\hline
\end{tabular}
\end{center}
\label{tab:split_upavia}
\end{table}

The performance of HSI classification of the proposed network is evaluated on three datasets: the Indian Pines dataset, the Salinas dataset, and the University of Pavia dataset, as shown in Figure~\ref{fig:hyp_img}.  The Indian Pines dataset consists of 145$\times$145 pixels and 220 spectral reflectance bands covering the range from 0.4 to 2.5 $\mu m$ with a spatial resolution of 20 $m$.  The Indian Pines dataset originally has 16 classes but we only use 8 classes with relatively large numbers of samples.  The Salinas dataset consists of 512$\times$217 pixels and 224 spectral bands.  It contains 16 classes and is characterized by a high spatial resolution of 3.7 $m$.  The University of Pavia dataset contains 610$\times$340 pixels with 103 spectral bands covering the spectral range from 0.43 to 0.86 $\mu m$ with a spatial resolution of 1.3 $m$.  9 classes are in the dataset.  For the Salinas dataset and the University of Pavia dataset, we use all classes because both datasets do not contain classes with a relatively small number of samples.

We compare the performance of the proposed network to the one reported in~\cite{WHuJS15} that used a different deep CNN architecture and RBF kernel-based SVM on the three hyperspectral datasets.  The deep CNN used in~\cite{WHuJS15} consists of two convolutional layers and two fully connected layers, which is much shallower than our proposed network with nine convolutional layers.  Currently, for the Indian Pines and University of Pavia datasets, an approach using diversified Deep Belief Networks (D-DBN) \cite{PZhongJSTAR17} provides higher HSI classification accuracy than that of the network in \cite{WHuJS15}.  We also use D-DBN as a baseline in this work.  For the Indian Pines dataset, we also use three types of neural networks evaluated in~\cite{WHuJS15}: a two layer fully connected neural network (Two-layer NN), a fully connected neural network with one hidden layer (Three-layer NN), and the classic LeNet-5~\cite{YLeCunNC1989}.

\begin{table*}[t]
\caption{Comparison of hyperspectral classification performance among the proposed network and the baselines on three datasets (in perceptage).  The best performance among 20 train/test partitions is shown in parentheses.  The best performance among all methods is indicated in bold font.}
\begin{center}
\rowcolors{0}{}{lightgray}
\setlength{\tabcolsep}{28pt}
\renewcommand{\arraystretch}{1.4}
\begin{tabular}{l|l|l|l}
\hline
\multirow{2}{*}{Method} & \multicolumn{3}{c}{Performance} \\\cline{2-4}
& Indian Pines & Salinas & University of Pavia \\\hline\hline
Two-layer NN~\cite{WHuJS15} & 86.49 & $\cdot$ & $\cdot$ \\
RBF-SVM~\cite{WHuJS15} & 87.60 & 91.66 & 90.52 \\
Three-layer NN~\cite{WHuJS15,YChenJSTAR14} & 87.93 & $\cdot$ & $\cdot$ \\
LeNet-5~\cite{WHuJS15,YLeCunNC1989} & 88.27 & $\cdot$ & $\cdot$ \\
Shallower CNN~\cite{WHuJS15} & 90.16 & 92.60 & 92.56 \\
D-DBN~\cite{PZhongJSTAR17} & 91.03 $\pm$ 0.12 & $\cdot$ & 93.11 $\pm$ 0.06 \\
The proposed network & {\bf 93.61 $\pm$ 0.56 (94.24)} & {\bf 95.07 $\pm$ 0.23 (95.42)} & {\bf 95.97 $\pm$ 0.46 (96.73)} \\\hline
\end{tabular}
\end{center}
\label{tab:comp}
\end{table*}

\begin{table*}[t]
\caption{Performance comparison of the proposed network in percentage w.r.t. varying widths (number of kernels in each layer).}
\vspace{-0.5cm}
\begin{center}
\rowcolors{0}{}{lightgray}
\setlength{\tabcolsep}{27.0pt}
\renewcommand{\arraystretch}{1.4}
\begin{tabular}{l|c|c|c|c}
\hline
Dataset & 64 & 128 & 192 & 256 \\
\hline\hline
Indian Pines & 80.38 $\pm$ 14.20 & {\bf 93.61 $\pm$ 0.56} & 93.47 $\pm$ 0.41 & 92.79 $\pm$ 0.81 \\\hline
Salinas & 91.35 $\pm~$ 3.62 & 93.60 $\pm$ 0.58 & {\bf 95.07 $\pm$ 0.23} & 94.10 $\pm$ 0.55 \\\hline
University of Pavia & 94.77 $\pm~$ 0.83 & {\bf 95.97 $\pm$ 0.46} & 95.86 $\pm$ 0.50 & 95.78 $\pm$ 0.52 \\
\hline
\end{tabular}
\end{center}
\label{tab:comp_width}
\end{table*}

For a fair comparison, we randomly select 200 samples from each class and use them as training samples as in~\cite{WHuJS15}.  The rest are used for testing the proposed network.  The selected classes and the numbers of training and test samples of the three datasets are listed in Tables~\ref{tab:split_indian_pine},~\ref{tab:split_salinas}, and~\ref{tab:split_upavia}.   In the literature on HSI classification, different train/test dataset partitions are used to evaluate their approaches.  Among them, our dataset partition using 200 training samples has two advantages in evaluating the proposed network; i) evaluation with this partition can verify our contribution, which is building a deeper and wider network with a relatively small number of training samples and ii) \cite{WHuJS15} using this partition can provide reasonable performance of relatively good baselines, such as RBF-SVM and the shallower CNN.  For all experiments, we perform the random train/test partition 20 times and report mean and stand deviation of overall classification accuracy (OA).  We have carried out all the experiments on Caffe framework~\cite{JYangqingACMMM14} with a Titan X GPU.

\begin{table}[t]
\caption{Training time (in second) of the proposed network w.r.t. varying widths (number of kernels in each layer).}
\vspace{-0.5cm}
\begin{center}
\rowcolors{0}{}{lightgray}
\setlength{\tabcolsep}{14.0pt}
\renewcommand{\arraystretch}{1.4}
\begin{tabular}{l|c|c|c|c}
\hline
Dataset & 64 & 128 & 192 & 256 \\
\hline\hline
Indian Pines & 351 & 482 & 576 & 738 \\\hline
Salinas & 428 & 598 & 696 & 896 \\\hline
University of Pavia & 349 & 474 & 597 & 751 \\
\hline
\end{tabular}
\end{center}
\label{tab:comp_width_cost}
\end{table}

\subsection{HSI Classification}
\label{ssec:hic}

Table~\ref{tab:comp} shows a performance comparison among the proposed network and baselines on the datasets.   
Hu et al.~\cite{WHuJS15} only reports a single instance of classification performance without indicating if the value is the best or mean accuracy of multiple evaluations.  The proposed network provided improved performance over all the baselines on all datasets.  The mean of classification performance of the proposed network is better than the best baseline classification performance by 2.58 \%, 2.47 \%, and 2.86 \% for the Indian Pines dataset, the Salinas dataset, and the University of Pavia dataset, respectively.  This performance enhancement was achieved mainly by building a deeper and wider network as well as jointly exploiting the spatio-spectral information of the hyperspectral data.  Residual learning also helped improve the performance by optimizing training efficiency on a relatively small number of samples.  The groundtruth map (left) and the classification map (right) obtained by the proposed network for all datasets are also shown in Figure~\ref{fig:qual}.  The classification map is drawn from one arbitrary train/test partition among 20.

\begin{table}[t]
\caption{Performance comparison of the proposed network in percentage w.r.t. varying depths (number of residual learning modules).}
\vspace{-0.5cm}
\begin{center}
\rowcolors{0}{}{lightgray}
\setlength{\tabcolsep}{7.0pt}
\renewcommand{\arraystretch}{1.4}
\begin{tabular}{l|c|c|c}
\hline
Dataset & 1 & 2 & 3 \\
\hline\hline
Indian Pines & 92.74 $\pm$ 0.69 & {\bf 93.61 $\pm$ 0.56} & 92.63 $\pm$ 0.84 \\\hline
Salinas & 94.06 $\pm$ 0.26 & {\bf 95.07 $\pm$ 0.23} & 94.01 $\pm$ 0.47 \\\hline
University of Pavia & 95.63 $\pm$ 0.50 & {\bf 95.97 $\pm$ 0.46} & 95.66 $\pm$ 0.59 \\
\hline
\end{tabular}
\end{center}
\label{tab:comp_depth}
\end{table}

\begin{table}[t]
\caption{Training time (in second) of the proposed network w.r.t. varying depths (number of residual learning modules).}
\vspace{-0.5cm}
\begin{center}
\rowcolors{0}{}{lightgray}
\setlength{\tabcolsep}{19.0pt}
\renewcommand{\arraystretch}{1.4}
\begin{tabular}{l|c|c|c}
\hline
Dataset & 1 & 2 & 3 \\
\hline\hline
Indian Pines & 431 & 482 & 549 \\\hline
Salinas & 616 & 696 & 777 \\\hline
University of Pavia & 426 & 474 & 544 \\
\hline
\end{tabular}
\end{center}
\label{tab:comp_depth_cost}
\end{table}

\begin{figure*}
\begin{center}
\includegraphics[trim = 0mm 0mm 0mm 0mm,width=\textwidth]{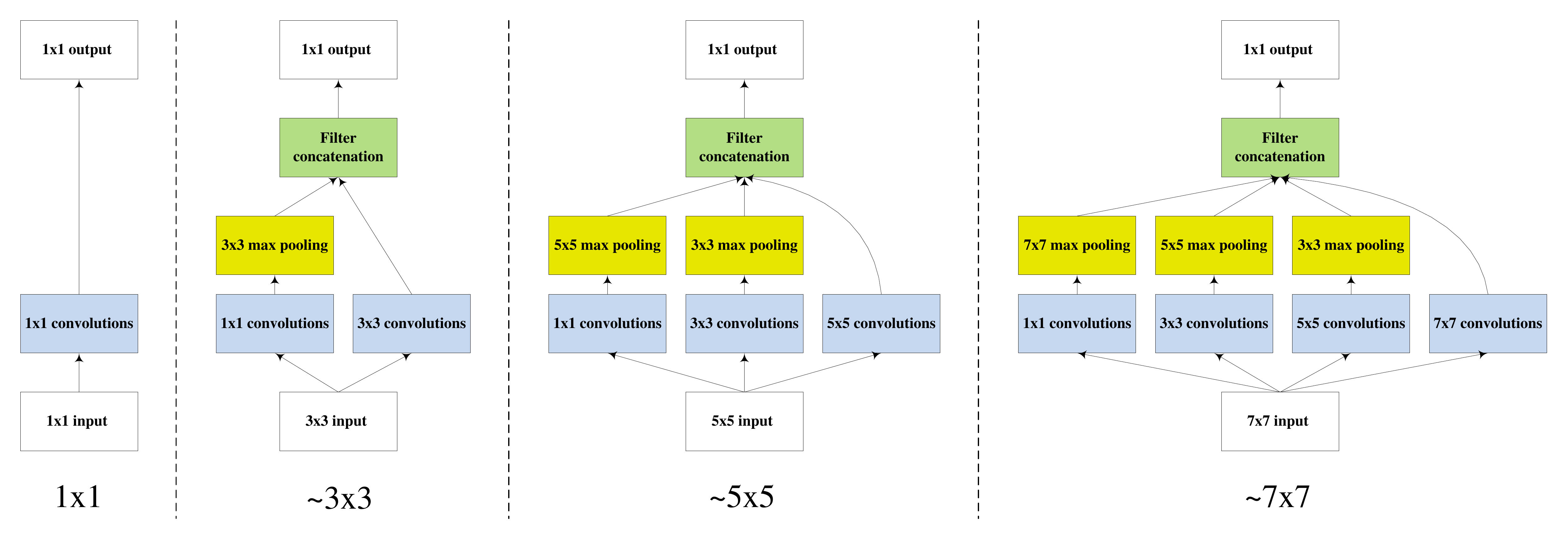}
\caption{Architecture of various multi-scale filter banks.}
\label{fig:architecture_various_inception}
\end{center}
\end{figure*}

\begin{table*}[t]
\caption{Performance comparison of the proposed network (in percentage) w.r.t. multi-scale filter banks with different configurations.  $\sim7\times 7$ means the multi-scale filter bank consisting of 1$\times$1, 3$\times$3, 5$\times$5, and 7$\times$7 convolution filters.}
\vspace{-0.5cm}
\begin{center}
\rowcolors{0}{}{lightgray}
\setlength{\tabcolsep}{27.0pt}
\renewcommand{\arraystretch}{1.4}
\begin{tabular}{l|c|c|c|c}
\hline
Dataset & 1$\times$1 & $\sim$3$\times$3 & $\sim$5$\times$5 & $\sim$7$\times$7 \\
\hline\hline
Indian Pines &  53.67 $\pm$ 16.63 & 87.37 $\pm$ 4.12 & {\bf 93.61 $\pm$ 0.56} & 93.47 $\pm$ 0.77 \\\hline
Salinas & 50.62 $\pm$ 30.87 & 92.08 $\pm$ 0.77 & {\bf 95.07 $\pm$ 0.23} & 94.20 $\pm$ 0.43 \\\hline
University of Pavia & 65.62 $\pm~$ 8.18 & 93.59 $\pm$ 1.35 & {\bf 95.97 $\pm$ 0.46} & 95.91 $\pm$ 0.50 \\
\hline
\end{tabular}
\end{center}
\label{tab:comp_inception}
\end{table*}

\subsection{Finding the Optimal Depth and Width of the Network}
\label{ssec:diff_depth_width}

To find the optimal width of the proposed network, we evaluate the network by varying the number of convolutional filters (i.e., the number of kernels): 64, 128, 192, and 256 for all three datasets.  Table~\ref{tab:comp_width} shows the performance of the proposed network with the varying numbers of kernels (network width) while Table~\ref{tab:comp_width_cost} shows training time for all cases.  For the Indian Pines dataset and the University of Pavia dataset, 128 is the optimal width for the best performance while 192 is the best one for the Salinas dataset.  Since the Salinas dataset contains more training samples from the larger number of classes than other datasets, more weights seem to be necessary to achieve optimal performance.  As shown in Table~\ref{tab:comp_width} and~\ref{tab:comp_width_cost}, adding more filters to the optimal network not only causes reduction in performance but also results in an increase in computational cost.

We also evaluate the proposed network with various depths in order to find the optimal depth.  Depth can be varied by using different numbers of residual learning modules.  Performance comparison of the proposed network with varying numbers of residual learning modules is shown in Table~\ref{tab:comp_depth}.  Table~\ref{tab:comp_depth_cost} shows training time for all cases.  For all the three datasets, using two residual learning modules achieves the best performance among all variations.  Using three residual learning modules may face an overfitting issue, which results in performance degradation.  It is also shown in Table~\ref{tab:comp_depth_cost} that using three residual learning modules turns out to be computationally very expensive.

On the basis of these evaluations, we choose the network with two residual learning modules and the width of 128 for each layer for both the Indian Pines dataset and the University of Pavia dataset.  For the Salinas dataset, the network with two residual learning modules and the width of 192 for each layer is selected.

\subsection{Effectiveness of the Multi-scale Filter Bank}
\label{ssec:perf_inception}

To verify the effectiveness of the multi-scale filter bank used to jointly exploit the spatio-temporal information together, we compare the proposed network to the network without the multi-scale filter bank, which use only a 1$\times$1 filter in the first layer.  We also compare to the network with the multi-scale filter bank with a different configuration: 1$\times$1, 3$\times$3, 5$\times$5, and 7$\times$7.  Figure~\ref{fig:architecture_various_inception} shows architectures of all various multi-scale filter banks.  As shown in Table~\ref{tab:comp_inception},  the multi-scale filter bank significantly outperforms the network without it (1x1 only) for all the three datasets (by 39.94 $\%$ for the Indian Pines dataset, 44.45 $\%$ for the Salinas dataset, and 30.35 $\%$ for the University of Pavia in mean classification performance).  The drastic performance degradation is mainly caused by two reasons; i) no joint exploitation of the spatio-spectral information is performed and ii) data augmentation by mirroring local regions cannot be used due to the non-existence of spatial filtering.

We also compare the proposed network to the one multi-scale filter banks with different configurations.  As shown in Table~\ref{tab:comp_inception}, The performance degradation from using the multi-scale filter bank with all the filters up to 7$\times$7 denoted by $\sim$7$\times$7 is caused by 'spillover' near class boundaries resulted from using the spatial filter of 7$\times$7.  Therefore, we choose to use a multi-scale filter bank with 1$\times$1, 3$\times$3, and 5$\times$5 for the proposed network.

\begin{figure*}[t]
\begin{center}
\begin{tabular}{ccc}
\subfloat[Indian Pines]{
\begin{tabular}{c}
\includegraphics[trim = 60mm 85mm 55mm 85mm,width=.28\textwidth]{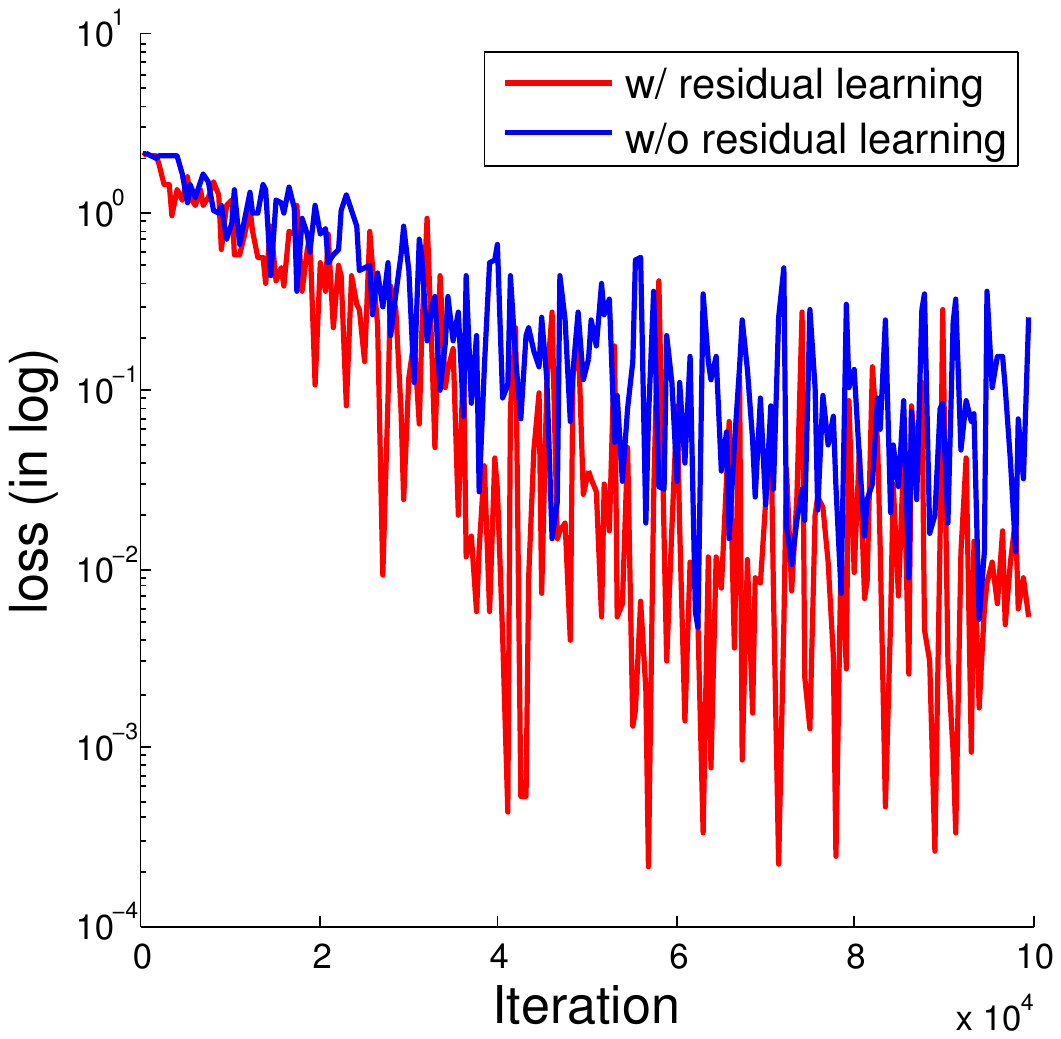}\\
\includegraphics[trim = 60mm 85mm 55mm 85mm,width=.28\textwidth]{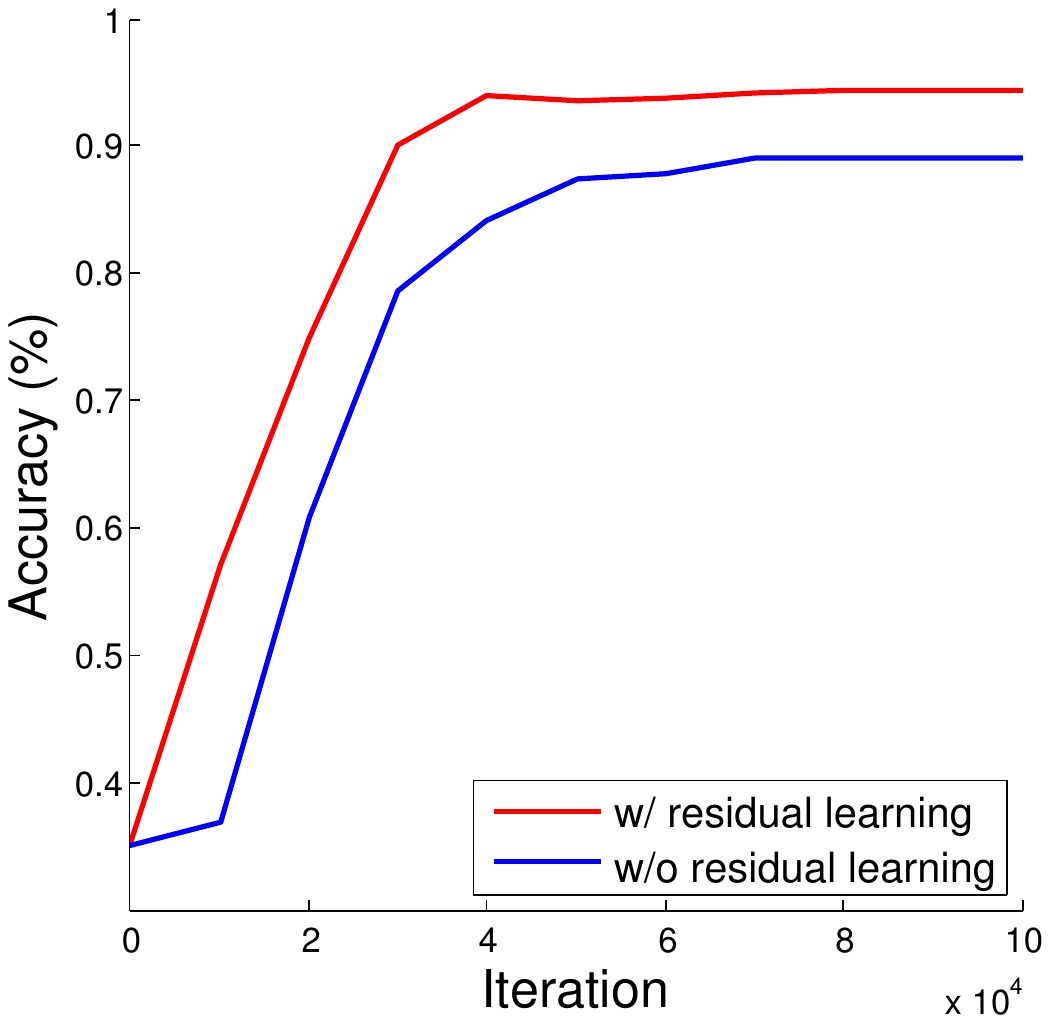}
\end{tabular}
\label{fig:eval_rl_indian_pines}} &
\subfloat[Salinas]{
\begin{tabular}{c}
\includegraphics[trim = 55mm 85mm 50mm 85mm,width=.305\textwidth]{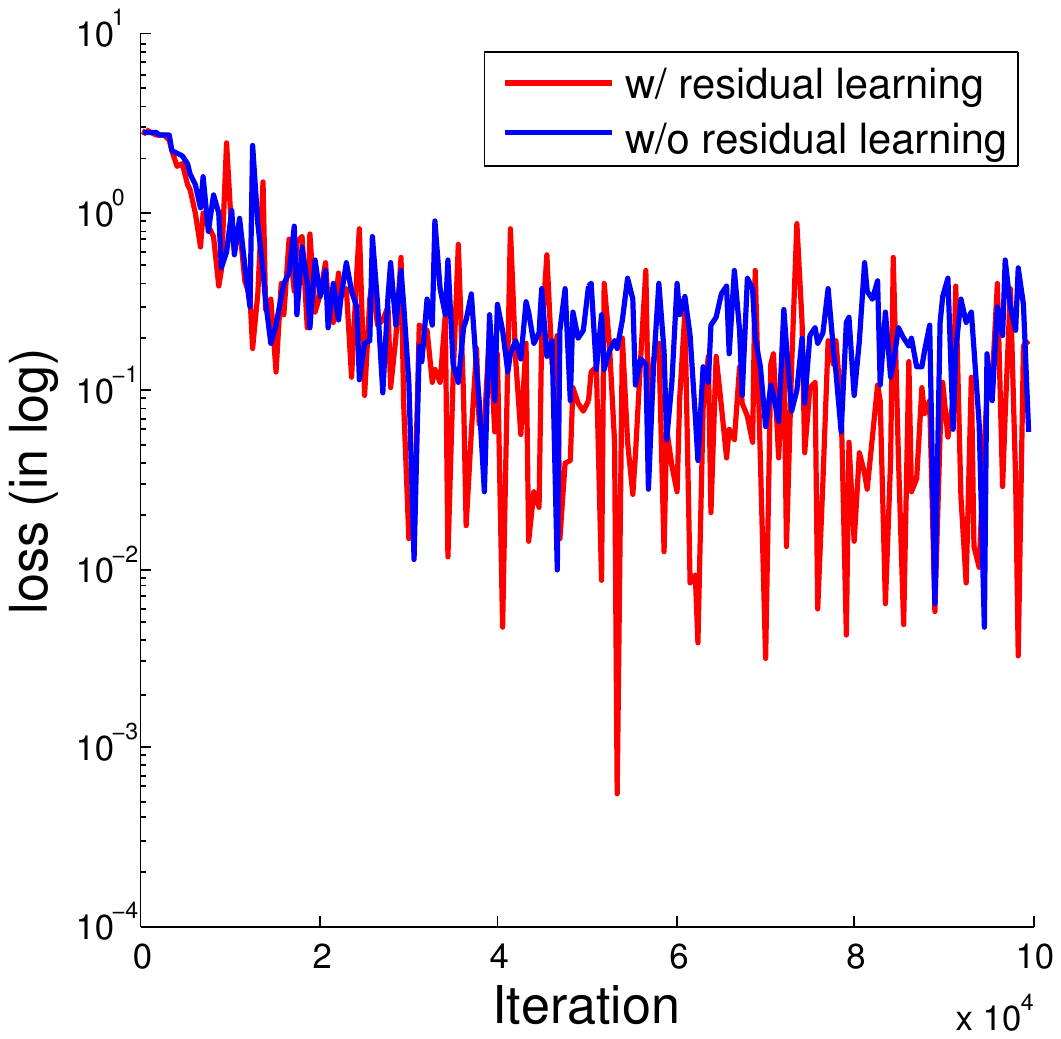}\\
\includegraphics[trim = 55mm 85mm 50mm 85mm,width=.305\textwidth]{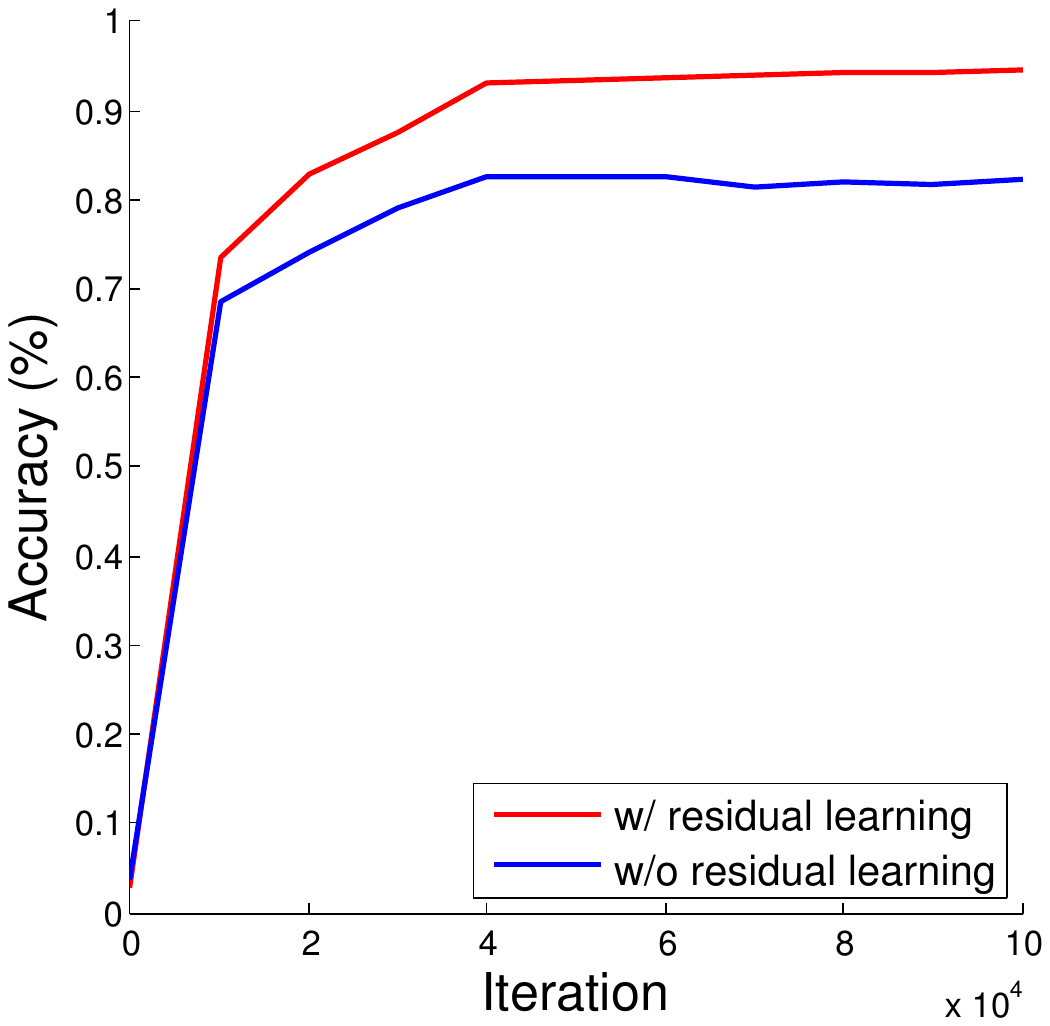}
\end{tabular}
\label{fig:eval_rl_salinas}}
\subfloat[University of Pavia]{
\begin{tabular}{c}
\includegraphics[trim = 55mm 85mm 55mm 85mm,width=.29\textwidth]{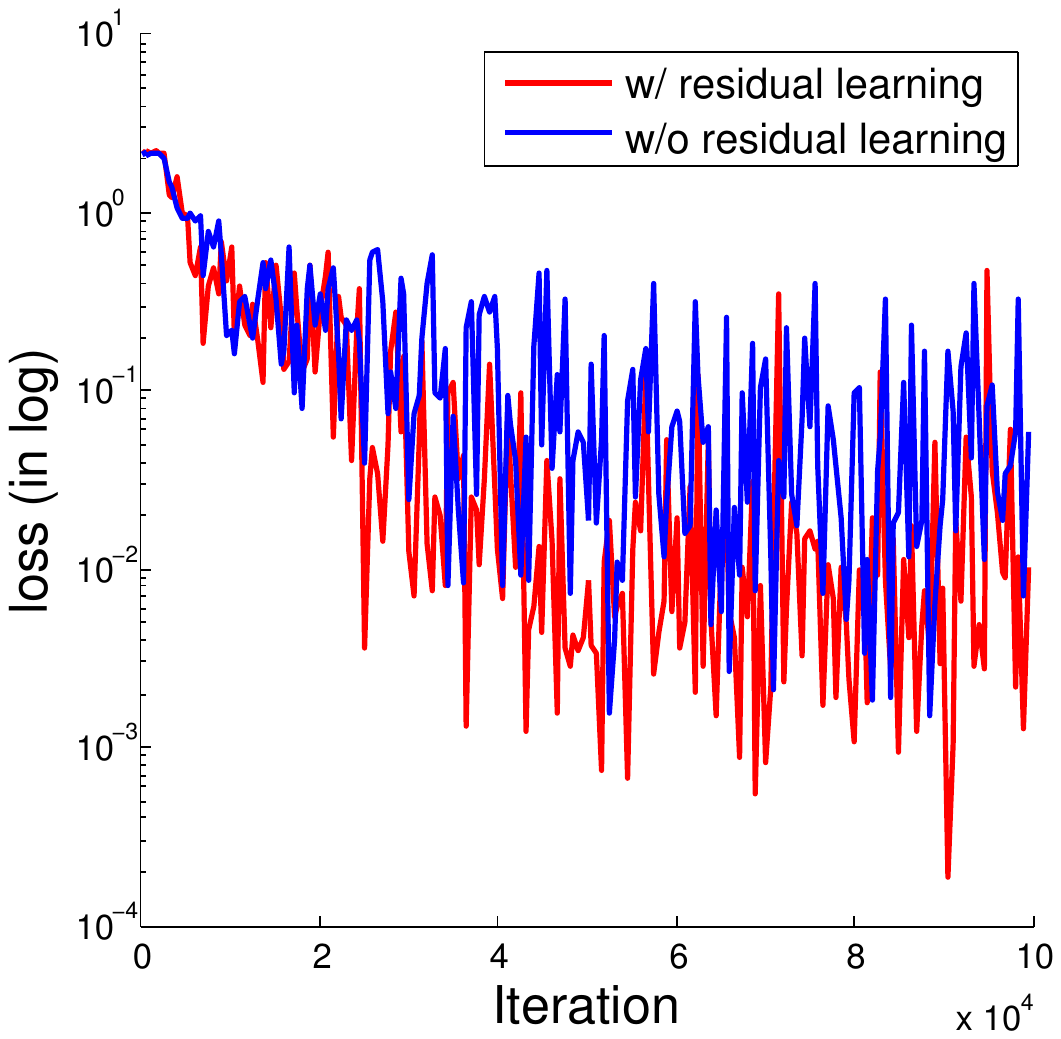}\\
\includegraphics[trim = 55mm 85mm 55mm 85mm,width=.29\textwidth]{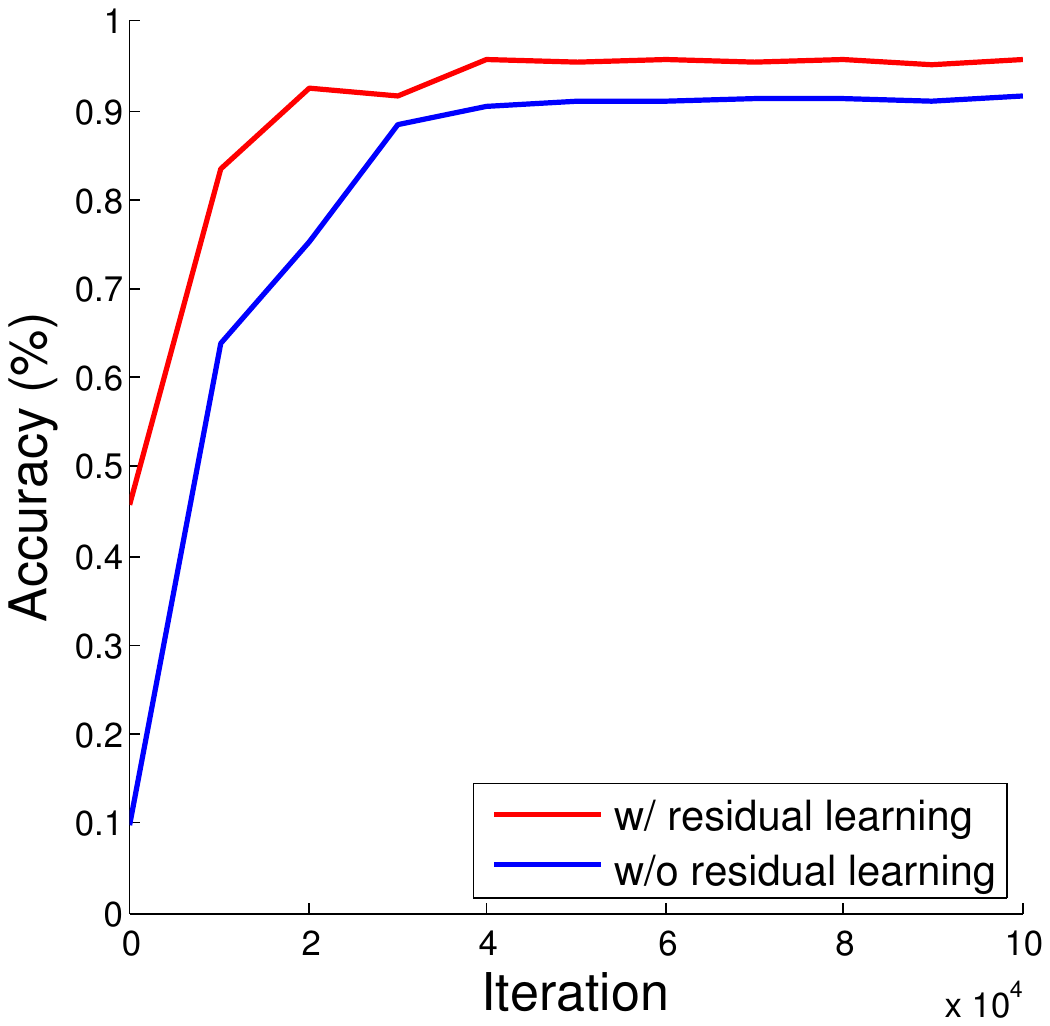}
\end{tabular}
\label{fig:eval_rl_paviau}}
\end{tabular}
\caption{{\bf Evaluation of effectiveness of residual learning.} Training loss (top) and classification accuracy (bottom) on three datasets with the proposed network and the network with the first residual learning module replaced with two convolutional layers are provided as a function of training iterations.  Note that `w/ residual learning' is the proposed architecture and `w/o residual learning' is the modified architecture replacing the first residual learning modules with regular two nonlinear layers as two sequential convolutional layers with the same nonlinear layers.}
\label{fig:eval_rl}
\end{center}
\end{figure*}

\subsection{Effectiveness of Residual Learning}
\label{ssec:perf_res_learn}

To verify the effectiveness of the ``residual learning'', we also compare the performance of the proposed network to a similar network with the first residual module replaced with regular two convolutional layers, as shown in Table~\ref{tab:comp_rl}.  Both the networks are built on the same number of convolutional layers, which is 9.  It was found that the network without using residual learning modules at all failed to converge in training due mainly to the small size training data.  The network with the first residual learning module replaced with two convolutional layers also failed to optimize the network parameters resulting in sub-optimal performance, as shown in Table~\ref{tab:comp_rl}.  Figure~\ref{fig:eval_rl} shows the comparison of training loss and classification accuracy as a function of training iterations for the two networks, which are calculated from one arbitrary train/test partition.  From the training loss in the plots of the first row of Figure~\ref{fig:eval_rl},  we observe that the proposed network achieves lower loss both during learning and at the end of the iterations than the other network. The second row of the Figure~\ref{fig:eval_rl} also shows that lower loss during learning leads to improved classification accuracy.  These observations support that residual learning greatly improves overall learning efficiency resulting in both lower training loss and higher classification accuracy.

\begin{table}[t]
\caption{Classification performance comparison of the proposed network and the network with the first residual learning module replaced with regular convolutional layers (in percentage).}
\vspace{-0.5cm}
\begin{center}
\rowcolors{0}{}{lightgray}
\setlength{\tabcolsep}{12.5pt}
\renewcommand{\arraystretch}{1.4}
\begin{tabular}{l|c|c}
\hline
Dataset & w/ conv. layer & w/ residual learning \\
\hline\hline
Indian Pines &  49.73 $\pm$ 24.58 & {\bf 93.61 $\pm$ 0.56} \\\hline
Salinas &  46.75 $\pm$ 25.98 & {\bf 95.07 $\pm$ 0.23} \\\hline
University of Pavia &  50.23 $\pm$ 27.78 & {\bf 95.97 $\pm$ 0.46} \\
\hline
\end{tabular}
\end{center}
\label{tab:comp_rl}
\end{table}

\subsection{Performance Changes according to Training Set Size }
\label{ssec:perf_diff_train_examples}

\begin{table*}[t]
\caption{Performance comparison of the proposed network (in percentage) w.r.t. the number of training examples per a class.}
\begin{center}
\rowcolors{0}{}{lightgray}
\setlength{\tabcolsep}{11.0pt}
\renewcommand{\arraystretch}{1.4}
\begin{tabular}{l|l|c|c|c|c|c}
\hline
\multicolumn{1}{c|}{Dataset} & \multicolumn{1}{c|}{Method} & 50 & 100 & 200 & 400 & 800 \\
\hline\hline
\multirow{2}{*}{Indian Pines} & MKL~\cite{YGuTGARS16} & 77.40 $\pm$ 1.78 & 80.63 $\pm$ 0.99 & $\cdot$ & $\cdot$ & $\cdot$ \\
& The proposed network & 80.50 $\pm$ 3.93 & 87.39 $\pm$ 0.88 & 93.61 $\pm$ 0.56 & 94.68 $\pm$ 0.47 & $\cdot$ \\\hline
\multirow{2}{*}{Salinas} & MKL~\cite{YGuTGARS16} & 89.33 $\pm$ 0.44 & 90.60 $\pm$ 0.43 & $\cdot$ & $\cdot$ & $\cdot$ \\
& The proposed network & 91.36 $\pm$ 1.11 & 93.15 $\pm$ 0.43 & 95.07 $\pm$ 0.23 & 96.55 $\pm$ 0.29 & 97.14 $\pm$ 0.53 \\\hline
\multirow{2}{*}{University of Pavia} & MKL~\cite{YGuTGARS16} & 91.52 $\pm$ 0.98 & 92.72 $\pm$ 0.33 & $\cdot$ & $\cdot$ & $\cdot$ \\
& The proposed network & 91.39 $\pm$ 0.80 & 93.10 $\pm$ 0.45 & 95.97 $\pm$ 0.46 & 96.81 $\pm$ 0.25 & 97.31 $\pm$ 0.26 \\
\hline
\end{tabular}
\end{center}
\label{tab:comp_inception}
\end{table*}

To analyze the effects of training dataset size in learning the proposed network, we compare the performance of the proposed network as the size of training dataset is changed: 50, 100, 200, 400, or 800 examples per a class.  Table~\ref{tab:comp_inception} presents classification accuracy of the proposed network w.r.t. training dataset size.  For the Indian Pines dataset, we do not perform learning with 800 examples per a class because several classes have insufficient examples (e.g. 483 for {\it Grass-pasture}, 478 for {\it Hay-windrowed}, 593 for {\it Soybean-clean}).

As expected, the classification accuracy of the proposed network monotonically increases as training dataset size increases.  We also note that even for smaller training dataset size, such as 50 and 100, the proposed network provides higher accuracy than multiple kernel learning (MKL)-based HSI classification~\cite{YGuTGARS16}, as shown in Table~\ref{tab:comp_inception}.

\begin{table*}[t]
\caption{{\bf Confusion matrix.} Groundtruth labels and classified classes are given along $x$ and $y$ axes, respectively.  The numbers along the axes correspond to the class numbers in Table~\ref{tab:split_indian_pine},~\ref{tab:split_salinas}, and~\ref{tab:split_upavia} for the three datasets, respectively.  Per a class, best accuracy is indicated by bold font.}
\vspace{-0.8cm}
\begin{center}
\begin{tabular}{c}
\subfloat[Indian Pines]{
\setlength{\tabcolsep}{16.3pt}
\renewcommand{\arraystretch}{1.2}
\begin{tabular}{r||c|c|c|c|c|c|c|c}
\hline
& 1 & 2 & 3 & 4 & 5 & 6 & 7 & 8 \\
\hline\hline
1 & {\bf 90.1} \% & 1.5 \% & 0.0 \% & 0.0 \% & 1.0 \% & 4.9 \% & 2.2 \% & 0.3 \% \\\hline
2 & 1.8 \% & {\bf 97.1} \% & 0.0 \% & 0.0 \% & 0.0 \% & 1.1 \% & 0.0 \% & 0.0 \% \\\hline
3 & 0.0 \% & 0.0 \% & {\bf 100.0} \% & 0.0 \% & 0.0 \% & 0.0 \% & 0.0 \% & 0.0 \% \\\hline
4 & 0.0 \% & 0.0 \% & 0.0 \% & {\bf 100.0} \% & 0.0 \% & 0.0 \% & 0.0 \% & 0.0 \% \\\hline
5 & 1.3 \% & 0.0 \% & 0.1 \% & 0.0 \% & {\bf 95.9} \% & 2.2 \% & 0.5 \% & 0.0 \% \\\hline
6 & 5.5 \% & 3.7 \% & 0.0 \% & 0.0 \% & 3.1 \% & {\bf 87.1} \% & 0.7 \% & 0.0 \% \\\hline
7 & 2.0 \% & 0.8 \% & 0.0 \% & 0.0 \% & 0.0 \% & 0.8 \% & {\bf 96.4} \% & 0.0 \% \\\hline
8 & 0.0 \% & 0.0 \% & 0.6 \% & 0.0 \% & 0.0 \% & 0.0 \% & 0.0 \% & {\bf 99.4} \% \\\hline
\end{tabular}}
\vspace{0.2cm}\\

\subfloat[Salinas]{
\setlength{\tabcolsep}{2.2pt}
\renewcommand{\arraystretch}{1.2}
\begin{tabular}{r||c|c|c|c|c|c|c|c|c|c|c|c|c|c|c|c}
\hline
& 1 & 2 & 3 & 4 & 5 & 6 & 7 & 8 & 9 & 10 & 11 & 12 & 13 & 14 & 15 & 16 \\
\hline\hline
1 & {\bf 100.0} \% & 0.0 \% & 0.0 \% & 0.0 \% & 0.0 \% & 0.0 \% & 0.0 \% & 0.0 \% & 0.0 \% & 0.0 \% & 0.0 \% & 0.0 \% & 0.0 \% & 0.0 \% & 0.0 \% & 0.0 \% \\\hline
2 & 0.0 \% & {\bf 100.0} \% & 0.0 \% & 0.0 \% & 0.0 \% & 0.0 \% & 0.0 \% & 0.0 \% & 0.0 \% & 0.0 \% & 0.0 \% & 0.0 \% & 0.0 \% & 0.0 \% & 0.0 \% & 0.0 \% \\\hline
3 & 0.0 \% & 0.0 \% & {\bf 100.0} \% & 0.0 \% & 0.0 \% & 0.0 \% & 0.0 \% & 0.0 \% & 0.0 \% & 0.0 \% & 0.0 \% & 0.0 \% & 0.0 \% & 0.0 \% & 0.0 \% & 0.0 \% \\\hline
4 & 0.0 \% & 0.0 \% & 0.0 \% & {\bf 99.3} \% & 0.7 \% & 0.0 \% & 0.0 \% & 0.0 \% & 0.0 \% & 0.0 \% & 0.0 \% & 0.0 \% & 0.0 \% & 0.0 \% & 0.0 \% & 0.0 \% \\\hline
5 & 0.0 \% & 0.0 \% & 0.0 \% & 0.5 \% & {\bf 98.5} \% & 0.0 \% & 0.0 \% & 0.0 \% & 0.0 \% & 0.2 \% & 0.0 \% & 0.2 \% & 0.0 \% & 0.0 \% & 0.6 \% & 0.0 \% \\\hline
6 & 0.0 \% & 0.0 \% & 0.0 \% & 0.0 \% & 0.0 \% & {\bf 100.0} \% & 0.0 \% & 0.0 \% & 0.0 \% & 0.0 \% & 0.0 \% & 0.0 \% & 0.0 \% & 0.0 \% & 0.0 \% & 0.0 \% \\\hline
7 & 0.2 \% & 0.0 \% & 0.0 \% & 0.0 \% & 0.0 \% & 0.0 \% & {\bf 99.8} \% & 0.0 \% & 0.0 \% & 0.0 \%  & 0.0 \% & 0.0 \% & 0.0 \% & 0.0 \% & 0.0 \% & 0.0 \% \\\hline
8 & 0.0 \% & 0.0 \% & 0.0 \% & 0.0 \% & 0.0 \% & 0.0 \% & 0.0 \% & {\bf 83.4} \% & 0.0 \% & 0.9 \% & 0.0 \%  & 0.0 \% & 0.0 \% & 0.3 \% & 15.5 \% & 0.0 \% \\\hline
9 & 0.0 \% & 0.0 \% & 0.0 \% & 0.0 \% & 0.0 \% & 0.0 \% & 0.0 \% & 0.0 \% & {\bf 99.6} \% & 0.0 \% & 0.4 \% & 0.0 \% & 0.0 \% & 0.0 \% & 0.0 \% & 0.0 \% \\\hline
10 & 0.0 \% & 0.0 \% & 1.0 \% & 0.0 \% & 0.0 \% & 0.2 \% & 0.0 \% & 0.3 \% & 0.3 \% & {\bf 94.6} \% & 1.6 \% & 1.0 \% & 0.0 \% & 0.6 \% & 0.4 \% & 0.0 \% \\\hline
11 & 0.0 \% & 0.0 \% & 0.0 \% & 0.0 \% & 0.0 \% & 0.0 \% & 0.0 \% & 0.0 \% & 0.0 \% & 0.0 \% & {\bf 99.3} \% & 0.7 \% & 0.0 \% & 0.0 \% & 0.0 \% & 0.0 \% \\\hline
12 & 0.0 \% & 0.0 \% & 0.0 \% & 0.0 \% & 0.0 \% & 0.0 \% & 0.0 \% & 0.0 \% & 0.0 \% & 0.0 \% & 0.0 \% & {\bf 100.0} \% & 0.0 \% & 0.0 \% & 0.0 \% & 0.0 \% \\\hline
13 & 0.0 \% & 0.0 \% & 0.0 \% & 0.0 \% & 0.0 \% & 0.0 \% & 0.0 \% & 0.0 \% & 0.0 \% & 0.0 \% & 0.0 \% & 0.0 \% & {\bf 100.0} \% & 0.0 \% & 0.0 \% & 0.0 \% \\\hline
14 & 0.0 \% & 0.0 \% & 0.0 \% & 0.0 \% & 0.0 \% & 0.0 \% & 0.0 \% & 0.0 \% & 0.0 \% & 0.0 \% & 0.0 \% & 0.0 \% & 0.0 \% & {\bf 100.0} \% & 0.0 \% & 0.0 \% \\\hline
15 & 0.0 \% & 0.0 \% & 0.0 \% & 0.0 \% & 0.0 \% & 0.0 \% & 0.0 \% & 0.0 \% & 0.0 \% & 0.0 \% & 0.0 \% & 0.0 \% & 0.0 \% & 0.0 \% & {\bf 100.0} \% & 0.0 \% \\\hline
16 & 0.0 \% & 0.0 \% & 0.0 \% & 0.1 \% & 0.1 \% & 0.0 \% & 0.5 \% & 1.0 \% & 0.0 \% & 0.0 \% & 0.0 \% & 0.0 \% & 0.0 \% & 0.2 \% & 0.1 \% & {\bf 98.0} \% \\\hline
\end{tabular}}
\vspace{0.2cm}\\

\subfloat[University of Pavia]{
\setlength{\tabcolsep}{13.7pt}
\renewcommand{\arraystretch}{1.2}
\begin{tabular}{r||c|c|c|c|c|c|c|c|c}
\hline
& 1 & 2 & 3 & 4 & 5 & 6 & 7 & 8 & 9 \\
\hline\hline
1 & {\bf 94.6} \% & 0.0 \% & 1.2 \% & 0.0 \% & 0.0 \% & 0.0 \% & 2.8 \% & 1.2 \% & 0.0 \& \\\hline
2 & 0.0 \% & {\bf 96.0} \% & 0.0 \% & 1.7 \% & 0.0 \% & 2.3 \% & 0.0 \% & 0.0 \% & 0.0 \% \\\hline
3 & 0.5 \% & 0.0 \% & {\bf 95.5} \% & 0.0 \% & 0.0 \% & 0.3 \% & 0.0 \% & 4.7 \% & 0.0 \% \\\hline
4 & 0.0 \% & 3.1 \% & 0.0 \% & {\bf 95.9} \% & 0.0 \% & 0.9 \% & 0.0 \% & 0.0 \% & 0.0 \% \\\hline
5 & 0.0 \% & 0.0 \% & 0.0 \% & 0.0 \% & {\bf 100.0} \% & 0.0 \% & 0.0 \% & 0.0 \% & 0.0 \% \\\hline
6 & 0.0 \% & 4.4 \% & 0.0 \% & 0.2 \% & 0.0 \% & {\bf 94.1} \% & 0.0 \% & 1.2 \% & 0.0 \% \\\hline
7 & 2.0 \% & 0.0 \% & 0.0 \% & 0.0 \% & 0.0 \% & 0.0 \% & {\bf 97.5} \% & 0.4 \% & 0.0 \% \\\hline
8 & 1.7 \% & 0.1 \% & 8.9 \% & 0.0 \% & 0.0 \% & 0.5 \% & 0.0 \% & {\bf 88.8} \% & 0.0 \% \\\hline
9 & 0.1 \% & 0.0 \% & 0.0 \% & 0.0 \% & 0.0 \% & 0.0 \% & 0.4 \% & 0.0 \% & {\bf 99.5} \% \\\hline
\end{tabular}}
\end{tabular}

\label{tab:conf_matrix}
\end{center}
\end{table*}

\begin{table*}[t]
\caption{Categorization of the false positives w.r.t. the pixel distance to the boundary}
\vspace{-0.5cm}
\begin{center}
\rowcolors{0}{}{lightgray}
\setlength{\tabcolsep}{18.5pt}
\renewcommand{\arraystretch}{1.4}
\begin{tabular}{l|ccc|ccc}
\hline
\multirow{2}{*}{Dataset} & \multicolumn{3}{c|}{\# of FP / \# of test data} & \multicolumn{3}{c}{Percentage} \\
\cline{2-7}
& 0 & 1 & $\geq$ 2 & 0 & 1 & $\geq$ 2 \\
\hline\hline
Indian pines & 93 / 717 & 80 / 109 & 310 / 5478 & 12.97 \% & 11.28 \% & 5.66 \% \\
\hline
Salinas & 94 / 1093 & 81 / 1082 & 2688 / 48754 & 8.60 \% & 7.49 \% & 5.51 \% \\
\hline
University of Pavia & 254 / 3455 & 299 / 4135 & 1737 / 33386 & 7.35 \% & 7.23 \% & 4.30 \% \\
\hline
\end{tabular}
\end{center}
\label{tab:fp_anal}
\end{table*}

\subsection{False Positives Analysis}
\label{ssec:FP_anal} 

Table~\ref{tab:conf_matrix} shows confusion matrices for three datasets, which are calculated from one arbitrary train/test partition.  For the Indian Pines dataset, the proposed network presents the performance below 95 $\%$ in only two classes that are {\it corn-notill} and {\it soybean-mintill}, among the eight classes.  As shown in the Table~\ref{tab:split_indian_pine}, the two classes are the ones with much larger numbers of samples than others.  The network learning with relatively small training data seems to fail to represent overall spectral characteristics of the classes.  Similarly, approximately $5 \%$ of false positives of each of the two classes are labeled as the other class because the spectral distributions of the two classes are more widespread than others.  Similar tendency is shown for the Salinas dataset.  The proposed network performed worst for the two classes with more test data, which are {\it grapes untrained} and {\it vineyard untrained}, as shown in Table~\ref{tab:split_salinas}: 83.4 $\%$ for {\it grapes untrained} and 89.4 $\%$ for {\it vineyard untrained}.  Most false positives from each of the two classes are the ones misclassified as the other class of the two classes.  For the University of Pavia dataset, the classification performance of the {\it bricks} class is noticeably worse, which is less than 90 $\%$.  Most false positives of the {\it bricks} class are classified as {\it gravels}.

To evaluate how the proposed network performs for pixels near boundaries between different classes, we categorized all the pixels according to the pixel distance to the boundary.  Pixels on the boundary are labelled as zero.  Similarly, pixels near boundary with one pixel apart are labelled as one.  The rest are labelled as $\geq$ 2.  Note that we use neighboring 5$\times$5 pixels for exploiting spatial information of each pixel.  For pixels labelled as $\geq$ 2, their 5$\times$5 neighboring pixels are from the same class.  Table~\ref{tab:fp_anal} shows the number of false positives versus all the test data within each pixel category for all the three datasets.  For all datasets, it is observed that larger portions of false positives are generated near boundaries as expected.  The false positives close to class boundaries are one of major factors for performance degradation of the proposed network.  The pixels far from the boundaries by more than one pixel distance are not affected by `spillover' and therefore less prone to misclassification.

\section{Conclusion}
\label{sec:concl}

In the proposed work, we have built a fully convolutional neural network with a total of 9 layers, which is much deeper than other existing convolutional networks for HSI classification. It is well known that a suitably optimized deeper network can in general lead to improved performance over shallower networks. To enhance the learning efficiency of the proposed network trained on a relatively sparse training samples a newly introduced learning approach called residual learning has been used.  To leverage both spectral and spatial information embedded in hyperspectral images, the proposed network jointly exploits local spatio-spectral interactions by using a multi-scale filter bank at the initial stage of the network. The multi-scale filter bank consists of three convolutional filters with different sizes: two filters ($3\times 3$ and $5\times 5$) are used to exploit local spatial correlations while $1\times 1$ is used to address spectral correlations.

As supported by the experimental results, the proposed network provided enhanced classification performance on the three benchmark datasets over current state-of-the-art approaches using different CNN architectures. The improved performance is mainly from i) using a deeper network with enhanced training and ii) joint exploitation of spatio-spectral information. The depth (the number of layers) and width (the number of kernels used in each layer) of the proposed network as well as the number of residual learning modules are determined by cross validation. The classification performance also shows that the proposed network with two residual learning modules outperforms the one with only one module, which supports the effectiveness of the residual learning incorporated into the proposed network.


%

\ifCLASSOPTIONcaptionsoff
  \newpage
\fi



\bibliographystyle{IEEEtran}
\bibliography{references.bib}

\begin{thebibliography}{10}
\providecommand{\url}[1]{#1}
\csname url@samestyle\endcsname
\providecommand{\newblock}{\relax}
\providecommand{\bibinfo}[2]{#2}
\providecommand{\BIBentrySTDinterwordspacing}{\spaceskip=0pt\relax}
\providecommand{\BIBentryALTinterwordstretchfactor}{4}
\providecommand{\BIBentryALTinterwordspacing}{\spaceskip=\fontdimen2\font plus
\BIBentryALTinterwordstretchfactor\fontdimen3\font minus
  \fontdimen4\font\relax}
\providecommand{\BIBforeignlanguage}[2]{{%
\expandafter\ifx\csname l@#1\endcsname\relax
\typeout{** WARNING: IEEEtran.bst: No hyphenation pattern has been}%
\typeout{** loaded for the language `#1'. Using the pattern for}%
\typeout{** the default language instead.}%
\else
\language=\csname l@#1\endcsname
\fi
#2}}
\providecommand{\BIBdecl}{\relax}
\BIBdecl

\bibitem{YChenJSTAR14}
Y.~Chen, Z.~Lin, X.~Zhao, G.~Wang, and Y.~Gu, ``Deep learning-based
  classification of hyperspectral data,'' \emph{IEEE Journal of Selected Topics
  in applied Earth Observations and Remote Sensing (J-STARS)}, vol.~7, no.~6,
  pp. 2094--2107, 2014.

\bibitem{WHuJS15}
W.~Hu, Y.~Huang, L.~Wei, F.~Zhang, and H.~Li, ``Deep convolutional neural
  networks for hyperspectral image classification,'' \emph{Journal of Sensors},
  vol. 2015.

\bibitem{WZhaoTGARS16}
W.~Zhao and S.~Du, ``Spectral-spatial feature extraction for hyperspectral
  image classification: A dimension reduction and deep learning approach,''
  \emph{IEEE Transactions on Geoscience and Remote Sensing (TGARS)}, vol.~54,
  no.~8, 2016.

\bibitem{YChenTGARS16}
Y.~Chen, H.~Jiang, C.~Li, X.~Jia, and P.~Ghamisi, ``Deep feature extraction and
  classification of hyperspectral images based on convolutional neural
  networks,'' \emph{IEEE Transactions on Geoscience and Remote Sensing
  (TGARS)}, vol.~54, no.~10, pp. 6232--6251, 2016.

\bibitem{PLiuJSTAR17}
P.~Liu, H.~Zhang, and K.~Eom, ``Active deep learning for classification of
  hyperspectral images,'' \emph{IEEE Journal of Selected Topics in applied
  Earth Observations and Remote Sensing (J-STARS)}, no.~10, pp. 712--724, 2017.

\bibitem{PZhongJSTAR17}
P.~Zhong, Z.~Gong, S.~Li, and C.-B. Sch{\'' o}nlieb, ``Learning to diversify
  deep belief networks for hyperspectral image classification,'' \emph{IEEE
  Journal of Selected Topics in applied Earth Observations and Remote Sensing
  (J-STARS)}, no.~99, pp. 1--15, 2017.

\bibitem{YChenJSTAR15}
Y.~Chen, X.~Zhao, and X.~Jia, ``Spectral-spatial classification of
  hyperspectral data based on deep belief network,'' \emph{IEEE Journal of
  Selected Topics in applied Earth Observations and Remote Sensing (J-STARS)},
  vol.~8, no.~6, pp. 2381--2392, 2015.

\bibitem{TLiICIP14}
T.~Li, J.~Zhang, and Y.~Zhang, ``Classification of hyperspectral image based on
  deep belief networks,'' in \emph{IEEE Conference on Image Processing (ICIP)},
  2014.

\bibitem{KPearsonPM1901}
K.~Pearson, ``On lines and planes of closest fit to systems of points in
  space,'' \emph{Philosophical Magazine}, vol.~2, no.~11, pp. 559--572, 1901.

\bibitem{XWangJSTAR17}
X.~Wang, Y.~Kong, Y.~Gao, and Y.~Cheng, ``Dimensionality reduction for
  hyperspectral data based on pairwise constraint discriminative analysis and
  nonnegative sparse divergence,'' \emph{IEEE Journal of Selected Topics in
  applied Earth Observations and Remote Sensing (J-STARS)}, no.~10, pp.
  1552--1562, 2017.

\bibitem{KHeCVPR16}
K.~He, X.~Zhang, S.~Ren, and J.~Sun, ``Deep residual learning for image
  recognition,'' in \emph{IEEE conference on Computer Vision and Pattern
  Recognition (CVPR)}, 2016.

\bibitem{CSzegedyCVPR15}
C.~Szegedy, W.~Liu, Y.~Jia, P.~Sermanet, S.~Reed, D.~Anguelov, D.~Erhan,
  V.~Vanhoucke, and A.~Rabinovich, ``Going deeper with convolutions,'' in
  \emph{IEEE conference on Computer Vision and Pattern Recognition (CVPR)},
  2015.

\bibitem{JLongCVPR15}
J.~Long, E.~Shelhamer, and T.~Darrell, ``Fully convolutional networks for
  semantic segmentation,'' in \emph{IEEE conference on Computer Vision and
  Pattern Recognition (CVPR)}, 2015.

\bibitem{HLeeIGARSS16}
H.~Lee and H.~Kwon, ``Contextual deep cnn based hyperspectral classification,''
  in \emph{IEEE International Geoscience and Remote Sensing Symposium
  (IGARSS)}, 2016.

\bibitem{YLeCunNC1989}
Y.~LeCun, B.~Boser, J.~S. Denker, D.~Henderson, R.~Howard, W.~Hubbard, and
  L.~Jackel, ``Backpropagation applied to handwritten zip code recognition,''
  \emph{Nerual Computation}, vol.~1, pp. 541--551, 1989.

\bibitem{AKrizhevskyNIPS12}
A.~Krizhevsky, I.~Sutskever, and G.~Hinton, ``Imagenet classification with deep
  convolutional neural networks,'' in \emph{Conference on Neural Information
  Processing Systems (NIPS)}, 2012.

\bibitem{JDengCVPR09}
J.~Deng, W.~Dong, L.~J.~J. R.~Socher, K.~Li, and L.~Fei-Fei, ``Imagenet: A
  large-scale hierarchical image database,'' in \emph{IEEE conference on
  Computer Vision and Pattern Recognition (CVPR)}, 2009.

\bibitem{KSimonyanICLR15}
K.~Simonyan and A.~Zisserman, ``Very deep convolutional networks for
  large-scale image recognition,'' in \emph{International Conference on
  Learning Representations (ICLR)}, 2015.

\bibitem{PGurramTGARS13}
P.~Gurram and H.~Kwon, ``Sparse kernel-based ensemble learning with fully
  optimized kernel parameters for hyperspectral classification problems,''
  \emph{IEEE Transactions on Geoscience and Remote Sensing (TGARS)}, vol.~51,
  pp. 787--802, 2013.

\bibitem{YGuTGARS16}
Y.~Gu, T.~Liu, X.~Jia, J.~A. Benediktsson, and J.~Chanussot, ``Nonlinear
  multiple kernel learning with multiple-structure-element extended
  morphological profiles for hyperspectral image classification,'' \emph{IEEE
  Transactions on Geoscience and Remote Sensing (TGARS)}, vol.~54, pp.
  3235--3247, 2016.

\bibitem{FMorsierTGARS16}
F.~de~Morsier, M.~Borgeaud, V.~Gass, J.-P. Thiran, and D.~Tuia, ``Kernel
  low-rank and sparse graph for unsupervised and semi-supervised classification
  of hyperspectral images,'' \emph{IEEE Transactions on Geoscience and Remote
  Sensing (TGARS)}, vol.~54, pp. 3410--3420, 2016.

\bibitem{JLiuTGARS16}
J.~Liu, Z.~Wu, J.~Li, A.~Plaza, and Y.~Yuan, ``Probabilistic-kernel
  collaborative representation for spatial-spectral hyperspectral image
  classification,'' \emph{IEEE Transactions on Geoscience and Remote Sensing
  (TGARS)}, vol.~54, pp. 2371--2384, 2016.

\bibitem{QWangTGARS16}
Q.~Wang, Y.~Gu, and D.~Tuia, ``Discriminative multiple kernel learning for
  hyperspectral image classification,'' \emph{IEEE Transactions on Geoscience
  and Remote Sensing (TGARS)}, vol.~54, pp. 3912--3927, 2016.

\bibitem{BGuoTIP08}
B.~Guo, S.~R. Gunn, R.~I. Demper, and J.~D.~B. Nelson, ``Customizing kernel
  functions for {SVM}-based hyperspectral image classification,'' \emph{IEEE
  Transactions on Image Processing (TIP)}, vol.~17, pp. 622--629, 2008.

\bibitem{LYangJSTAR17}
L.~Yang, M.~Wang, S.~Yang, R.~Zhang, and P.~Zhang, ``Sparse spatio-spectral
  lap{SVM} with semisupervised kernel propagation for hyperspectral image
  classification,'' \emph{IEEE Journal of Selected Topics in applied Earth
  Observations and Remote Sensing (J-STARS)}, no.~99, pp. 1--9, 2017.

\bibitem{RRoscherTGARS16}
R.~Roscher and B.~Waske, ``Shapelet-based sparse representation for landcover
  classification of hyperspectral images,'' \emph{IEEE Transactions on
  Geoscience and Remote Sensing (TGARS)}, vol.~54, pp. 1623--1634, 2016.

\bibitem{JinlinLiuTGARS16}
J.~Liu and W.~Lu, ``A probabilistic framework for spectral-spatial
  classification of hyperspectral images,'' \emph{IEEE Transactions on
  Geoscience and Remote Sensing (TGARS)}, vol.~54, pp. 5375--5384, 2016.

\bibitem{AZhetabianTGARS16}
A.~Zehtabian and H.~Ghassemian, ``Automatic object-based hyperspectral image
  classification using complex diffusions and a new distance metric,''
  \emph{IEEE Transactions on Geoscience and Remote Sensing (TGARS)}, vol.~54,
  pp. 4106--4114, 2016.

\bibitem{SJiaTGARS16}
S.~Jia, J.~Hu, Y.~Xie, L.~Shen, X.~Jia, and Q.~Li, ``Gabor cube selection based
  multitask joint sparse representation for hyperspectral image
  classification,'' \emph{IEEE Transactions on Geoscience and Remote Sensing
  (TGARS)}, vol.~54, pp. 3174--3187, 2016.

\bibitem{JunshiXiaTGARS16}
J.~Xia, J.~Chanussot, P.~Du, and X.~He, ``Rotation-based support vector machine
  ensemble in classification of hyperspectral data with limited training
  samples,'' \emph{IEEE Transactions on Geoscience and Remote Sensing (TGARS)},
  vol.~54, pp. 1519--1531, 2016.

\bibitem{ZZhongTGARS16}
Z.~Zhong, B.~Fan, K.~Ding, H.~Li, S.~Xiang, and C.~Pan, ``Efficient multple
  feature fusion with hashing for hyperspectral imagery classification: A
  comparative study,'' \emph{IEEE Transactions on Geoscience and Remote Sensing
  (TGARS)}, vol.~54, pp. 4461--4478, 2016.

\bibitem{JXiaTGARS16}
J.~Xia, L.~Bombrun, T.~Adali, Y.~Berthoumieu, and C.~Germain,
  ``Spectral-spatial classification of hyperspectral images using ica and
  edge-preserving filter via an ensemble strategy,'' \emph{IEEE Transactions on
  Geoscience and Remote Sensing (TGARS)}, vol.~54, pp. 4971--4982, 2016.

\bibitem{HYangTGARS16}
H.~Yang and M.~Crawford, ``Spectral and spatial proximity-based manifold
  alignment for multitemporal hyperspectral image classification,'' \emph{IEEE
  Transactions on Geoscience and Remote Sensing (TGARS)}, vol.~54, pp. 51--64,
  2016.

\bibitem{MToksozTGARS16}
M.~Toks{\" o}z and {\' I}.~Ulusoy, ``Hyperspectral image classification via
  basic thresholding classifier,'' \emph{IEEE Transactions on Geoscience and
  Remote Sensing (TGARS)}, vol.~54, pp. 4039--4051, 2016.

\bibitem{PZhongTIP10}
P.~Zhong and R.~Wang, ``Learning conditional random fields for classification
  of hyperspectral images,'' \emph{IEEE Transactions on Image Processing
  (TIP)}, vol.~19, pp. 1890--1907, 2010.

\bibitem{KBernardTIP12}
K.~Bernard, Y.~Tarabaika, J.~Angulo, J.~Chanussot, and J.~A. Benediktsson,
  ``Spectral-spatial classification of hyperspectral data based on a stochastic
  minimum spanning forest approach,'' \emph{IEEE Transactions on Image
  Processing (TIP)}, vol.~21, pp. 2008--2021, 2012.

\bibitem{YGaoTIP14}
Y.~Gao, R.~Ji, P.~Cui, Q.~Dai, and G.~Hua, ``Hyperspectral image classification
  through bilayer graph-based learning,'' \emph{IEEE Transactions on Image
  Processing (TIP)}, vol.~23, pp. 2769--2778, 2014.

\bibitem{MBrellTGARS17}
M.~Brell, K.~Segl, L.~Guanter, and B.~Bookhagen, ``Hyperspectral and lidar
  intensity data fusion: A framework for the rigorous correction of
  illumination, anisotropic effects, and cross calibration,'' \emph{IEEE
  Transactions on Geoscience and Remote Sensing (TGARS)}, vol.~55, pp.
  2799--2810, 2017.

\bibitem{SJiaTGARS17a}
S.~Jia, J.~Hu, J.~Zhu, X.~gJia, and Q.~Li, ``Three-dimensional local binary
  patterns for hyperspectral imagery classification,'' \emph{IEEE Transactions
  on Geoscience and Remote Sensing (TGARS)}, vol.~55, pp. 2399--2413, 2017.

\bibitem{SJiaTGARS17b}
S.~Jia, B.~Deng, J.~Zhu, and Q.~Li, ``Superpixel-based multitask learning
  framework for hyperspectral image classification,'' \emph{IEEE Transactions
  on Geoscience and Remote Sensing (TGARS)}, vol.~55, pp. 2575--2588, 2017.

\bibitem{SMeiJSTAR17}
S.~Mei, Q.~Bi, J.~Ji, J.~Hou, and Q.~Du, ``Hyperspectral image classification
  by exploring low-rank property in spectral or/and spatial domain,''
  \emph{IEEE Journal of Selected Topics in applied Earth Observations and
  Remote Sensing (J-STARS)}, no.~99, pp. 1--12, 2017.

\bibitem{HSuJSTAR17}
H.~Su, Y.~Cai, and Q.~Du, ``Firefly-algorithm-inspired framework with band
  selection and extreme learning machine for hyperspectral image
  classification,'' \emph{IEEE Journal of Selected Topics in applied Earth
  Observations and Remote Sensing (J-STARS)}, no.~10, pp. 309--320, 2017.

\bibitem{EStroblICMLA13}
E.~Strobl and S.~Visweswaran, ``Deep multiple kernel learning,'' in \emph{IEEE
  International Conference on Machine Learning and Applications (ICMLA)}, 2013.

\bibitem{JYangqingACMMM14}
Y.~Jia*, E.~Shelhamer*, J.~Donahue, S.~Karayev, J.~Long, R.~Girshick,
  S.~Guadarrama, and T.~Darrell, ``Caffe: Convolutional architecture for fast
  feature embedding,'' in \emph{ACM Multimedia (ACMMM)}, 2014.

\end{thebibliography}
%



%

\begin{IEEEbiography}[{\includegraphics[width=1in,height=1.25in,clip,keepaspectratio,trim=17mm 0mm 17mm 0mm]{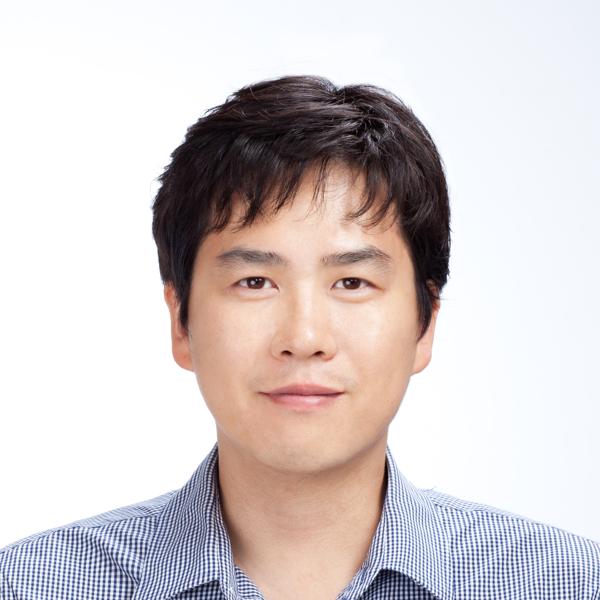}}]{Dr. Hyungtae Lee}
received the BS degree in electrical engineering and mechanical engineering from Sogang University, Seoul, Korea in 2006, MS degree from Korea Advanced Institute of Science and Technology (KAIST), Deajoen, Korea in 2008, and PhD degree from University of Maryland, College Park, MD, USA in 2014.  He works as a electrical engineering senior consultant for Booz Allen Hamilton Inc. at U.S. Army Research Laboratory in Adelphi, MD.  His current research interests include object, action, event, and pose recognition in computer vision, and machine learning.
\end{IEEEbiography}

\begin{IEEEbiography}[{\includegraphics[width=1in,height=1.25in,clip,keepaspectratio,trim=5mm 0mm 5mm 0mm]{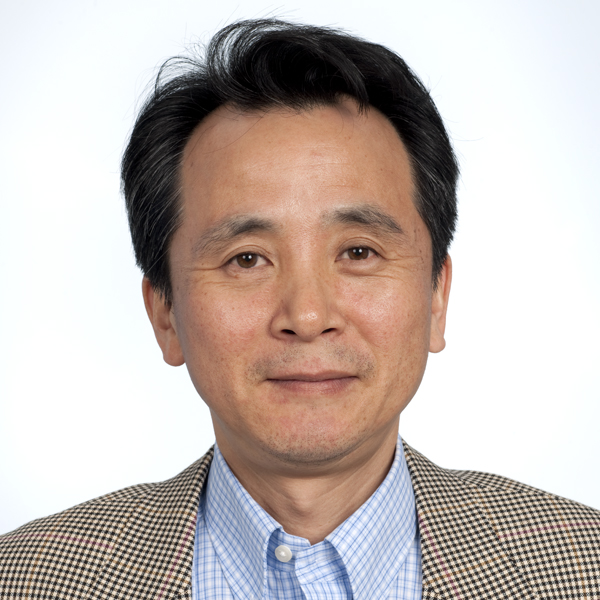}}]{Dr. Heesung Kwon} received the B.Sc. degree in Electronic Engineering from Sogang University, Seoul, Korea, in 1984, and the MS and Ph.D. degrees in Electrical Engineering from the State University of New York at Buffalo in 1995 and 1999, respectively. From 1983 to 1993, he was with Samsung Electronics Corp., where he worked as a senior research engineer. He was with the U.S. Army Research Laboratory (ARL), Adelphi, MD from 1996 to 2006 working on automatic target detection and hyperspectral signal processing applications. From 2006 to 2007, he was with Johns Hopkins University Applied Physics Laboratory (JHU/APL) working on biological standoff detection problems. Dr. Kwon rejoined ARL in August, 2007 as a senior electronics engineer, leading hyperspectral research efforts in the Image Processing Branch. Dr. Kwon is currently Associate Editor of IEEE Trans. on Aerospace and Electronic Systems. He also served as Lead Guest Editor of the Special Issue on Algorithms for Multispectral and Hyperspectral Image Analysis of the Journal of Electrical and Computer Engineering.  His current research interests include image/video analytics, human-autonomy interaction, hyperspectral signal processing, machine learning, and statistical learning. He has published over 100 journal, book chapters, and conference papers on these topics.
\end{IEEEbiography}






\end{document}